\documentclass[11pt]{article}
\usepackage[preprint]{acl}
\usepackage{microtype}

\usepackage{fontspec}
\usepackage[english,bidi=default]{babel}
\babelfont{rm}[Extension=.otf,UprightFont=*-regular,BoldFont=*-bold,
  ItalicFont=*-italic,BoldItalicFont=*-bolditalic]{texgyretermes}

\usepackage{amsmath,amssymb,amsthm}
\usepackage{booktabs}
\usepackage{multirow}
\usepackage{array}
\usepackage{graphicx}
\usepackage{xcolor}
\usepackage{tikz}
\usepackage{pgfplots}
\pgfplotsset{compat=1.17}
\usepgfplotslibrary{groupplots}
\usetikzlibrary{arrows.meta,positioning,calc,decorations.pathreplacing}

\pgfplotsset{
  paper/.style={
    tick label style={font=\scriptsize},
    label style={font=\footnotesize},
    title style={font=\footnotesize\bfseries},
    axis line style={black!55}, tick style={black!55},
    grid style={black!10},
    legend cell align=left,
    legend style={font=\scriptsize,draw=black!30,fill=white,fill opacity=0.92,text opacity=1,
                  inner sep=1.5pt,row sep=-1.5pt,line width=0.3pt},
  },
  est/.style={only marks,mark=*,mark size=1.9pt,color=#1,mark options={fill=#1,draw=#1},
              error bars/.cd,x dir=both,x explicit,
              error bar style={color=#1,line width=0.9pt},error mark options={rotate=90,mark size=1.6pt,line width=0.6pt}},
  seedpts/.style={only marks,mark=o,mark size=1.3pt,color=black!45,mark options={line width=0.35pt}},
}
\newcommand{\rowval}[2]{\node[anchor=east,font=\scriptsize,inner sep=0pt] at (axis cs:17.6,#1) {$#2$};}

\newif\ifethok
\ethokfalse
\IfFontExistsTF{Noto Sans Ethiopic}{%
   \newfontfamily\ethfont{Noto Sans Ethiopic}[Scale=0.90]\ethoktrue
}{%
   \IfFontExistsTF{NotoSansEthiopic-Regular.ttf}{%
      \newfontfamily\ethfont{NotoSansEthiopic-Regular.ttf}[Path=./,
          BoldFont=NotoSansEthiopic-Regular.ttf,ItalicFont=NotoSansEthiopic-Regular.ttf,
          Scale=0.90]\ethoktrue
   }{\ethokfalse}%
}
\newcommand{\schwa}{ə}   
\ifethok
  \newcommand{\gz}[2]{{\ethfont #1}}                    
  \newcommand{\gzt}[2]{{\ethfont #1}~(\textit{#2})}     
\else
  \newcommand{\gz}[2]{\textit{#2}}
  \newcommand{\gzt}[2]{\textit{#2}}
\fi

\theoremstyle{plain}
\newtheorem{theorem}{Theorem}
\newtheorem{lemma}{Lemma}
\newtheorem{proposition}{Proposition}

\newcommand{\Rt}{\mathcal{R}}
\newcommand{\fert}{\phi}
\newcommand{\Tok}{T}
\newcommand{\row}{\mathrm{row}}
\newcommand{\ord}{\mathrm{ord}}
\newcommand{\cons}{\mathrm{cons}}
\newcommand{\rpt}{\textup{\textsc{RootPat}}}
\newcommand{\chrf}{chrF\raisebox{0.2ex}{\small ++}}
\newcommand{\arm}[1]{\textsf{#1}}
\newcommand{\Dg}{\Delta_{\mathrm{gran}}}
\newcommand{\Dr}{\Delta_{\mathrm{root}}}
\newcommand{\Dread}{\Delta_{\mathrm{read}}}
\newcommand{\Dnd}{\Delta_{\mathrm{nd}}}
\newcommand{\Hcv}{\hat H_{\mathrm{cv}}}
\newcommand{\pholm}{p_{\mathrm{Holm}}}
\newcommand{\pseed}{p_{\mathrm{seed}}}

\definecolor{tokcol}{RGB}{31,101,163}
\definecolor{enccol}{RGB}{176,86,20}
\definecolor{datcol}{RGB}{50,120,60}
\definecolor{ctlcol}{RGB}{140,40,120}
\definecolor{lightbg}{RGB}{244,246,250}
\colorlet{totcol}{black!72}
\newcommand{\tokbox}[2]{\tikz[baseline=(n.base)]\node[draw=#1,fill=lightbg,rounded corners=1.5pt,
  inner xsep=2.2pt,inner ysep=1.6pt,font=\scriptsize,line width=0.5pt](n){#2};}

\title{What Do Tokenizer Comparisons Measure? Disentangling Granularity,\\
Root Assignment, and Word-Level Readout across Amharic and Hausa}

\author{
  \textbf{Abdul Rehman Antall}$^{\spadesuit}$\qquad
  \textbf{Baidya Nath Saha}$^{\clubsuit}$\\[3pt]
  $^{\spadesuit}$National University of Computer and Emerging Sciences (FAST-NUCES), Lahore, Pakistan\\
  $^{\clubsuit}$Concordia University of Edmonton, Alberta, Canada\\[2pt]
  \texttt{l230605@lhr.nu.edu.pk}\qquad\texttt{baidya.saha@concordia.ab.ca}
}

\begin{document}
\maketitle

\begin{abstract}
Morphology-aware tokenizers are expected to help morphologically rich languages, and their
reported gains over byte-pair encoding (BPE) are usually credited to their linguistic units. A comparison with BPE cannot support this attribution, since such a tokenizer also
segments text more finely (\emph{granularity}) and, on token-level tasks, changes how much the
sub-token \emph{readout} matters. We separate both factors from the effect of
grouping words by their roots (\emph{root assignment}) in Amharic, a Semitic language whose Ge'ez
abugida hides consonantal roots inside syllabic glyphs, with Latin-script Hausa as an orthographic
contrast. Our design combines a lossless root-and-pattern tokenizer with controls that change one
factor at a time, in two independent implementations. In both, real root assignment improves Amharic named-entity recognition over scrambled
roots ($+4.9$ and $+4.7$ F1), and root-frequency structure explains little of this gain. In small
from-scratch models, the readout choice moves F1 by up to $13.4$ points and reverses the
sign of every comparison of the root-and-pattern arms with BPE, whereas in two pretrained encoders it moves
F1 by less than one point. The granularity effect changes sign across languages
and implementations. We recommend decomposing tokenizer effects, and reporting the readout, before
a gain is attributed to linguistic structure.
\end{abstract}
\section{Introduction}
\label{sec:intro}

Sub-word tokenization is the first modelling decision in an NLP system, and it is often inherited
from an earlier system rather than chosen \citep{alqahtani2026stop}. The standard algorithms,
byte-pair encoding (BPE; \citealp{sennrich2016bpe}) and the unigram language model
\citep{kudo2018unigram}, derive their units from corpus statistics alone. The resulting
segmentation determines sequence length, cost and downstream quality, all of which vary widely
across languages \citep{rust2021tokenizer,ahia2023cost,petrov2023unfairness}. Language technology
also serves only a small fraction of the world's languages well, and data volume alone does not
explain this gap \citep{joshi2020state}. Large multilingual systems now cover many African
languages \citep{nllb2022,ogueji2021afriberta}, but they inherit a pipeline designed around English
and typologically similar languages. A recurring proposal in computational morphology is therefore
to build linguistic units, such as roots and affixes, into the tokenizer itself. Such
morphology-aware tokenizers are usually justified by a downstream comparison with BPE. This paper
asks what that comparison measures and under which conditions linguistic structure is actually
responsible for a gain.

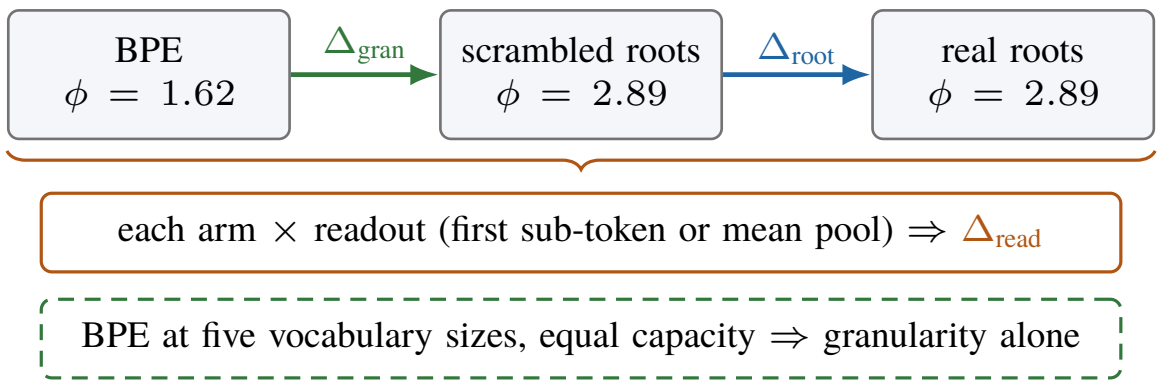
\begin{figure}[t]
\centering
{\footnotesize
\setlength{\tabcolsep}{3.5pt}
\begin{tabular}{@{}lll@{}}
\toprule
\textbf{Word} & \textbf{Real roots} & \textbf{Scrambled roots} \\
\midrule
\gz{ሰበረ}{s\"abb\"ar\"a} \textit{s\"abb\"ar\"a} &
  \tokbox{tokcol}{$\sqrt{\text{s-b-r}}$}\,\tokbox{black!45}{\"a\,\"a\,\"a} &
  \tokbox{ctlcol}{root 58}\,\tokbox{black!45}{\"a\,\"a\,\"a} \\[2pt]
\gz{ሰባሪ}{s\"abari} \textit{s\"abari} &
  \tokbox{tokcol}{$\sqrt{\text{s-b-r}}$}\,\tokbox{black!45}{\"a\,a\,i} &
  \tokbox{ctlcol}{root 203}\,\tokbox{black!45}{\"a\,a\,i} \\
\bottomrule
\end{tabular}}

\vspace{4pt}
\begin{tikzpicture}[
  box/.style={draw=black!55,rounded corners=2pt,align=center,minimum height=8.5mm,text width=1.72cm,
              font=\scriptsize,fill=lightbg,line width=0.5pt,inner sep=2pt},
  wide/.style={rounded corners=2pt,align=center,text width=7.0cm,minimum height=5.2mm,
               font=\scriptsize,line width=0.6pt,inner sep=2pt},
  ar/.style={-{Latex[length=1.9mm]},line width=0.9pt},
  lab/.style={font=\scriptsize,inner sep=1.5pt}]
\node[box] (b) {BPE\\ $\fert=1.62$};
\node[box,right=0.98cm of b] (s) {scrambled roots\\ $\fert=2.89$};
\node[box,right=0.98cm of s] (r) {real roots\\ $\fert=2.89$};
\draw[ar,datcol] (b) -- node[above,lab,text=datcol]{$\Dg$} (s);
\draw[ar,tokcol] (s) -- node[above,lab,text=tokcol]{$\Dr$} (r);
\draw[enccol,line width=0.6pt,decorate,decoration={brace,mirror,amplitude=3.5pt}]
  ([yshift=-2pt]b.south west) -- ([yshift=-2pt]r.south east);
\node[wide,draw=enccol,below=0.34cm of s] (rd)
  {each arm $\times$ readout (first sub-token or mean pool) $\Rightarrow$ {\color{enccol}$\Dread$}};
\node[wide,draw=datcol,densely dashed,below=0.16cm of rd] (sw)
  {BPE at five vocabulary sizes, equal capacity $\Rightarrow$ granularity alone};
\end{tikzpicture}
\caption{Top: two forms of $\sqrt{s\text{-}b\text{-}r}$ share only their first fidel but receive
the same root token, next to a template token of vowel orders; the scrambled control keeps the
templates and randomises the root identifier (identifiers illustrative). Bottom: a comparison with
BPE changes $\Dg$, $\Dr$ and $\Dread$ together, and each control changes one of them. $\fert$:
tokens per word (Study~I, Amharic NER).}
\label{fig:lattice}
\end{figure}

Amharic, a Semitic language of the Afroasiatic family, is a demanding case for statistical
segmentation. It builds many words by root-and-pattern morphology: a consonantal root, typically
triliteral, is interleaved with a vocalic template that marks tense, aspect, voice and derivation
\citep{mccarthy1981prosodic,leslau1995amharic}. The root $\sqrt{s\text{-}b\text{-}r}$ produces
\gzt{ሰበረ}{s\"abb\"ar\"a} `he broke', \gzt{ተሰበረ}{t\"as\"abb\"ar\"a} `it was broken' and
\gzt{ሰባሪ}{s\"abari} `breaker'. Amharic is written in the Ge'ez abugida, in which each glyph
(\emph{fidel}) fuses a consonant with a vowel \citep{unicode15}. As a result, the forms of a
paradigm do not share their root as a substring: no contiguous piece of \gz{ሰበረ}{s\"abb\"ar\"a}
and \gz{ሰባሪ}{s\"abari} spells $\sqrt{s\text{-}b\text{-}r}$ in both (Figure~\ref{fig:lattice};
Theorem~\ref{thm:opacity}, Appendix~\ref{app:rootpat}). Any tokenizer whose units are character
substrings inherits this limitation at every vocabulary size. Hausa, a Chadic language of the same
family, provides a contrast: its Latin orthography writes consonants as separate characters, and
its morphology is largely concatenative \citep{newman2000hausa}.

These facts motivate a widely held hypothesis: tokenizers informed by linguistic structure should
outperform statistical ones on morphologically complex languages
\citep{klein2020life,amrhein2021nonconcatenative,gazit2025splinter,teklehaymanot2025movoc}. A
comparison of one morphology-aware tokenizer with BPE can neither confirm nor refute this
hypothesis, because such a tokenizer changes three things at once.
First, it segments more finely (\emph{granularity}). Second, it places boundaries such that words
sharing a root share a token (\emph{root assignment}). Third, on token-level tasks such as
named-entity recognition (NER), a word label must be read from several sub-token states, and the
additional sub-tokens change how much the choice of \emph{readout} matters. The three factors have
different consequences for practice. Granularity requires no linguistic resources, since any
tokenizer can be made finer by shrinking its vocabulary. Root assignment requires analysers or
annotated paradigms, which exist for few languages. Readout is an implementation choice that papers
often leave unreported, and an effect that originates there is not a property of the language at
all.

We separate the three factors with controls that change one at a time, in a deliberately narrow
setting. The main evidence comes from Amharic NER, a task in which all three factors can be
manipulated. Hausa has no root resources and contributes an orthographic contrast for granularity
and readout. Every control changes the vocabulary of the tokenizer, whereas a pretrained encoder
fixes its tokenizer, and the controls cannot be run behind one. The main comparisons therefore use
small Transformers trained from scratch, and we test separately whether the readout effect persists
in two pretrained encoders. To guard against artefacts of a single implementation, we ran the
central comparisons twice (Study~I and Study~II), with independent root analysers, code and
training set-ups, and we state as findings only the results that held in both studies. Our
contributions are as follows.

\begin{itemize}
\item \textbf{A methodology for tokenizer comparisons} (Section~\ref{sec:theory}): trained
      references that change one factor at a time, namely a scrambled-root control, a
      frequency-preserving permutation of root assignments, a capacity-matched granularity sweep
      and a readout crossed with every tokenizer. A lossless root-and-pattern factorisation of the
      Ge'ez abugida makes the root controls possible.
\item \textbf{Three findings that held in both implementations} (Section~\ref{sec:results}). Root
      assignment improves Amharic NER beyond a matched control. Readout choice is the largest
      single effect in from-scratch models and reverses tokenizer comparisons, whereas its effect
      in two pretrained encoders is small. The granularity component changes sign across languages
      and implementations.
\item \textbf{Reporting that separates ``not detected'' from ``negligible''}
      (Section~\ref{sec:res-repl}): equivalence verdicts at margins fixed before the runs,
      seed-level tests beside sentence-level intervals, and a ledger of every failed prediction.
\end{itemize}

\noindent For researchers who design tokenizers for morphologically rich languages, the practical
consequence is that a gain over BPE is evidence for linguistic structure only after granularity and
readout have been controlled.
\section{Related work}
\label{sec:related}

\paragraph{Sub-word tokenization and morphological alignment.}
BPE \citep{sennrich2016bpe} and unigram-LM segmentation \citep{kudo2018unigram}, commonly
implemented with SentencePiece \citep{kudo2018sentencepiece}, select units to compress a corpus
under a vocabulary budget; \citet{mielke2021between} survey the design space. Tokenizer quality
affects downstream performance \citep{rust2021tokenizer,bostrom2020bpe}, but the compression
rationale itself has been questioned: \citet{schmidt2024compression} find that minimising token
count does not improve downstream performance, and \citet{zouhar2023noiseless} propose R\'enyi
efficiency as a better predictor. Whether tokens should coincide with morphemes is equally
unsettled. For Telugu, \citet{vemula2025rethinking} find that unigram tokenizers outperform BPE in
most settings, with morphological alignment correlating only moderately with downstream
performance. Across 70 languages, \citet{arnett2025morphscore} find that morphological alignment
explains little of the variance in downstream performance, and \citet{alqahtani2026stop} call for
standardised, transparently reported tokenizer evaluation.

\paragraph{Morphology-aware segmentation for nonconcatenative languages.}
Unsupervised morphological segmentation \citep{creutz2007morfessor} has long been proposed as an
alternative to frequency-driven merges, with mixed results. \citet{saleva2021morphology} found no
consistent advantage over BPE for low-resource machine translation (MT), whereas derivationally
informed segmentation helps pretrained models interpret complex words
\citep{hofmann2021superbizarre}, and MorphBPE constrains merges with morpheme-boundary supervision
\citep{asgari2025morphbpe}. Nonconcatenative morphology remains difficult for both sub-word and
character models \citep{amrhein2021nonconcatenative}. For Semitic languages,
\citet{klein2020life} argued that linear word pieces cannot represent fused Hebrew morphology, and
\citet{gazit2025splinter} rearranged nonconcatenative text to let a standard tokenizer find
meaningful contiguous segments. Closest in design to our factorisation, \citet{hatch2025sre}
encode Semitic words as distinct root and template-stem tokens and report an average gain of $1.15$
BLEU over a BPE baseline; their comparison with tokens built from non-linguistic root and template
structures is the nearest precedent for our scrambled-root control. In the opposite direction, \citet{alakeel2026morphemes} find for Arabic
that the morphological alignment of a tokenizer is neither necessary nor sufficient for productive
root-and-pattern generation in large language models. For Ge'ez-script languages,
\citet{teklehaymanot2025movoc} integrate supervised morphological analysis into a BPE vocabulary
and obtain consistent improvements in intrinsic morpheme-boundary measures without significant MT
gains. Amharic MT has also benefited from transliteration \citep{gezmu2021amharic} and from
homophone normalisation \citep{belay2022normalization}, which our pipeline applies. Our controls
add to this literature a separation of root identity from root-frequency
structure, from segmentation granularity and from the readout, on a token-level task.

\paragraph{Word-level labelling with sub-word models.}
Token classification with sub-word models conventionally labels the first sub-token of each word
\citep{devlin2019bert}. For mBERT and XLM-RoBERTa across nine languages, \citet{acs2021pooling}
showed that this choice is the weakest of several pooling strategies for morphological tasks, and
\citet{dang2025tokenization} find that last-token pooling gives higher morphological probing
accuracy than averaging in mT5 and ByT5. The best aggregation thus depends on the model. We show that, in models trained
from scratch, the first-sub-token penalty can be large enough to reverse the direction of tokenizer
comparisons, whereas it nearly vanishes in two pretrained encoders.
\section{Methodology}
\label{sec:theory}

Our methodology replaces the comparison of two tokenizers with a set of trained reference
tokenizers, each of which differs from its neighbour in one factor only
(Figure~\ref{fig:lattice}; proofs in Appendix~\ref{app:proofs}).

\subsection{Decomposing a tokenizer comparison}
\label{sec:decomp}

Let $S(X)$ be the expected test score of a model trained with tokenizer $X$, and let the
\emph{fertility} of $X$ be its mean number of tokens per word. Comparing a morphology-aware
tokenizer $T$ with a baseline $B$ measures the total effect $S(T)-S(B)$. We split this total with a
\emph{reference} tokenizer $T^{\circ}$, which has the vocabulary and fertility of $T$ but assigns
roots without regard to morphology:
\begin{equation}
\begin{aligned}
S(T)-S(B) ={}& \underbrace{S(T^{\circ})-S(B)}_{\Dg\text{: granularity}}\\
 &+\underbrace{S(T)-S(T^{\circ})}_{\Dr\text{: root assignment}} .
\end{aligned}
\label{eq:decomp}
\end{equation}
The reference has to be trained. A study that trains only $B$ and $T$ observes two numbers, and
every split of their difference into a granularity part and a root part is consistent with these
two numbers (Proposition~\ref{prop:nonid}). A two-condition comparison therefore cannot tell
whether a gain comes from linguistic structure.

\subsection{Controls for root assignment}
\label{sec:controls}

All root controls keep the prefix, template and suffix tokens of every word and replace only its
root token. They consequently share the token inventory, the number of tokens of every word, the
fertility, the parameter count and the token positions of the real tokenizer
(Proposition~\ref{prop:scramble}).

The \emph{scrambled} control gives every analysable word type a root identifier drawn uniformly at
random. Root identifiers then carry about the same number of word types each, whereas real roots
are shared very unevenly, with many forms of frequent verbs and few of rare ones.

The \emph{permuted} control shuffles the real root assignment among word types. Every root
identifier keeps its exact number of word types, which preserves the skewed frequency profile of
real roots while destroying the correspondence between words and roots. A
\emph{stratified} variant shuffles only within deciles of corpus frequency, with the result that
frequent words keep frequent identifiers.

Real minus scrambled measures root assignment as defined in Equation~\ref{eq:decomp}. The
scrambled and permuted controls differ only in the frequency profile of root identifiers, and
their difference therefore measures how much of a root effect that profile alone could explain.

\subsection{A neutral reference for granularity}

To vary granularity without any linguistic information, we train BPE at five vocabulary sizes per
language, chosen to reach about $1.0$, $1.3$, $1.6$, $2.0$ and $2.3$--$2.4$ times the baseline
fertility. Every embedding table is padded to the size of the largest. All models of a language
thus have the same number of parameters and differ only in how finely they segment.

\subsection{Readout}
\label{sec:readout}

On a token-level task, the label of a word must be read from its sub-token states. We cross every
tokenizer with two readouts: the state of the first sub-token, which is the common default
\citep{devlin2019bert}, and the mean of all sub-token states of the word. The \emph{readout gain}
of a tokenizer $X$ is $\Gamma(X)=S_{\mathrm{pool}}(X)-S_{\mathrm{first}}(X)$, and the part of a
comparison between $T$ and $B$ that is due to the readout alone is
\begin{equation}
\Dread=\Gamma(T)-\Gamma(B).
\label{eq:readdecomp}
\end{equation}

A simple model of sub-token states shows why the readout gain should depend on the tokenizer.
Consider a word split into $m$ pieces, and let each sub-token state be the representation the
classifier needs, plus a fixed offset of that fragment and noise with variance $\sigma^2$ and
correlation $\varrho$ between fragments. The mean squared error of first-sub-token readout then
exceeds that of pooled readout by
\begin{equation}
\big(\beta_1^2-\bar\beta^{\,2}\big)+\sigma^2(1-\varrho)\,\tfrac{m-1}{m},
\label{eq:readout}
\end{equation}
with $\beta_1$ the offset of the first fragment and $\bar\beta$ the mean offset
(Proposition~\ref{thm:readout}). The first term compares bias, and the second compares variance.
The expression suggests three regimes. If fragments are unbiased, pooling helps, increasingly as
words are split into more pieces. If the first sub-token already represents the word, pooling
hurts; we expect this regime for Hausa, whose BPE units are close to whole words. If sub-tokens are
strongly contextualised, the offsets become similar and $\varrho$ approaches $1$, in which case the
readout matters little at any fertility; we expect this regime for pretrained encoders. We
formulated this account after Study~I, and Study~II tests its later predictions.

\subsection{A root-and-pattern tokenizer}
\label{sec:tok}

A root control requires a tokenizer in which roots are explicit. Each fidel of the core Ethiopic
Unicode block decomposes exactly into a consonant row and a vowel order,
$\row(c)=\lfloor(u(c)-\texttt{0x1200})/8\rfloor$ and $\ord(c)=(u(c)-\texttt{0x1200})\bmod 8$ for
code point $u(c)$. On this decomposition, the \rpt{} tokenizer maps an analysable word to at most
four tokens, $\langle\textsc{pfx}\rangle\,\langle\textsc{root}{:}\hat\Rt\rangle\,
\langle\textsc{tmpl}{:}\hat\pi\rangle\,\langle\textsc{sfx}\rangle$. The root token holds the
consonantal root, and the template token records the vowel order of each radical together with
every non-root unit of the stem. Words without an analysis receive a sub-word segmentation with
byte fallback. The factorisation is lossless up to homophone normalisation
(Theorem~\ref{thm:lossless}), a property we verify on every NER word type. Each factored word emits
separate root and template tokens, which prevents the tokenizer from being more compact than BPE at
the same vocabulary size: in our analyser arm, $71\%$ of word tokens are factored, and fertility is
therefore at least $1.71$ against $1.62$ for BPE (measured: $2.89$).

\subsection{Replication and statistics}
\label{sec:stats}

Study~I uses HornMorpho \citep{gasser2011hornmorpho} as root source and covers the full design.
Study~II re-implements this design with a different root analyser, new code and new training
set-ups, and adds the permutation controls and two pretrained encoders. The two studies are
compared only in the sign of each contrast, its size relative to seed variability, and its verdict;
their numbers are never pooled.

Differences are tested with a paired bootstrap over test sentences ($2{,}000$ resamples), which
recomputes the corpus metric from per-sentence counts
\citep{koehn2004significance,efron1993bootstrap}. Holm's correction is applied within pre-specified
comparison families \citep{holm1979simple,dror2018hitchhiker}. The bootstrap captures variation
over test sentences only, and we therefore report the between-seed standard deviation (SD)
separately and, for the central Study~II contrasts, a paired $t$-test across seeds ($\pseed$).
Each contrast receives one of three verdicts: \emph{effect} (Holm-adjusted point null rejected),
\emph{equivalent} (the $90\%$ bootstrap interval lies inside $(-\delta,\delta)$, the
two-one-sided-tests rule at $\alpha=0.05$; \citealp{lakens2017equivalence}) or \emph{inconclusive}.
Margins ($\delta\in\{1,2\}$ F1, $0.5$ \chrf{}) and comparison families were fixed in the analysis
code before the runs. In Study~II, a run with a development score of zero is rejected before
aggregation.
\section{Experiments}
\label{sec:results}

\subsection{Experimental setup}
\label{sec:setup}

\paragraph{Data.} NER uses the MasakhaNER~1.0 news data \citep{adelani2021masakhaner} for both
languages, since version 2.0 has no Amharic split. The data have nine BIO labels over
\textsc{per}, \textsc{loc}, \textsc{org} and \textsc{date}, with $1{,}750/250/500$ train/dev/test
sentences for Amharic and $1{,}912/276/552$ for Hausa. For MT, we translate English into Amharic
and Hausa on filtered OPUS-100 data \citep{zhang2020opus100}, with $20{,}000/1{,}000/1{,}000$ pairs
per language in Study~I, drawn from corpora matched in register ($44\%$ and $43\%$ religious
text). Paradigms
come from UniMorph \citep{batsuren2022unimorph} ($2{,}461$ Amharic lemmas), and morphological
analyses come from HornMorpho. Hausa has neither resource, and root-based arms consequently exist
only for Amharic.

\paragraph{Tokenizer arms.} The Amharic baselines are BPE and unigram (fertility $1.62$ and $1.89$
on NER). Three \rpt{} arms (fertility $2.82$--$2.90$) share their vocabulary budget ($5{,}387$
against $5{,}388$ types) and differ only in the source of their roots. The \emph{induced} arm needs
no analyser: its roots are the longest common consonantal subsequence across each UniMorph paradigm
($1{,}608$ roots). The \emph{lexicon} arm aligns words against the $1{,}703$ verb roots of
HornMorpho. The \emph{analyser} arm runs HornMorpho on every type and covers $80.4\%$ of NER types.
The scrambled control keeps the segmentation and fertility of the analyser arm exactly. A
fertility-matched BPE control reaches the same fertility only at a vocabulary of $600$. The
granularity sweep reaches fertility ratios of $1.00$--$2.42$ for Amharic and $1.00$--$2.27$ for
Hausa (details in Appendix~\ref{app:rootpat}).

\paragraph{Models.} Study~I trains small pre-LayerNorm Transformers from scratch
\citep{vaswani2017attention}: a two-layer NER encoder ($d=128$, about $0.97$M parameters) and a
2+2-layer MT encoder--decoder ($7.5$--$8.8$M parameters; Appendix~\ref{app:details}).
Non-embedding parameter counts are identical across NER tokenizer arms, sequence budgets equalise
words rather than sub-tokens, and all arms share the positional table. NER is scored by entity-level micro-F1 over words (reported $\times100$). MT is
scored by \chrf{} \citep{popovic2017chrf}, checked against sacreBLEU \citep{post2018sacrebleu},
with greedy decoding. Every configuration uses five seeds and is evaluated at the final epoch.

\paragraph{Study II.} The re-implementation keeps the MasakhaNER splits and the \rpt{} algorithms
of Study~I and changes the rest (Table~\ref{tab:studyii}, Appendix~\ref{app:details}). Its root
analyser is induced
from paradigm lists alone, without HornMorpho ($1{,}238$ roots). This analyser covers $67.2\%$ of
NER types, which lowers the fertility of the \rpt{} arms to $2.17$ (BPE: $1.53$). The NER encoder
is larger ($1.95$M parameters) and is selected on the development set. Study~II also fine-tunes two
pretrained encoders with three seeds each, XLM-R base and AfroXLM-R base
\citep{conneau2020xlmr,alabi2022afroxlmr}, whose own tokenizers have fertility $2.20$ on Amharic
NER.

In the following subsections, intervals are unadjusted $95\%$ paired-bootstrap intervals, and
significance statements use Holm-adjusted $p$-values. The underlying tables are in
Appendix~\ref{app:results}.

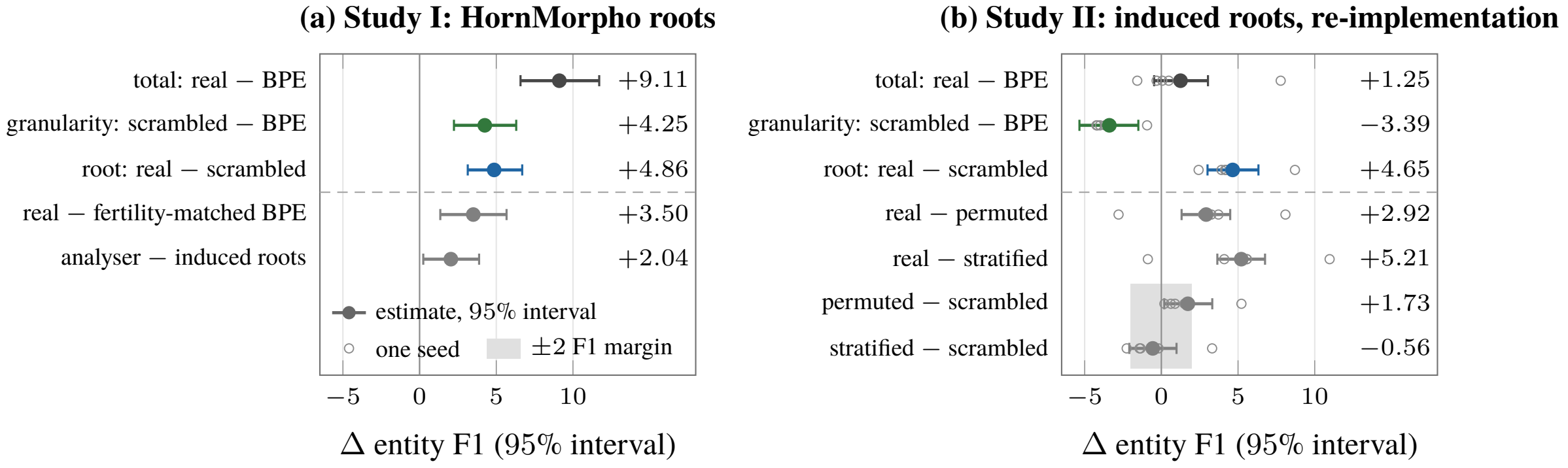
\begin{figure*}[t]
\centering
\begin{tikzpicture}
\begin{groupplot}[
  group style={group size=2 by 1, horizontal sep=3.7cm},
  paper, scale only axis, width=3.8cm, height=3.25cm,
  xmin=-6.5, xmax=18, xtick={-5,0,5,10},
  ymin=0.4, ymax=7.6, y dir=reverse, ytick={1,...,7},
  xmajorgrids, ymajorgrids=false, ytick style={draw=none},
  yticklabel style={font=\scriptsize},
  title style={yshift=-4pt},
  xlabel={$\Delta$ entity F1 ($95\%$ interval)},
]
\nextgroupplot[title={(a) Study I: HornMorpho roots}, ytick={1,...,5},
  yticklabels={{total: real $-$ BPE},{granularity: scrambled $-$ BPE},
   {root: real $-$ scrambled},{real $-$ fertility-matched BPE},{analyser $-$ induced roots}}]
\draw[black!45,line width=0.4pt] (axis cs:0,0.4) -- (axis cs:0,7.6);
\draw[black!35,densely dashed,line width=0.4pt] (axis cs:-6.5,3.5) -- (axis cs:18,3.5);
\addplot[est=totcol] coordinates {(9.11,1) += (2.60,0) -= (2.53,0)};
\addplot[est=datcol] coordinates {(4.25,2) += (2.05,0) -= (2.01,0)};
\addplot[est=tokcol] coordinates {(4.86,3) += (1.83,0) -= (1.72,0)};
\addplot[est=black!50] coordinates {(3.50,4) += (2.17,0) -= (2.15,0)
                                     (2.04,5) += (1.84,0) -= (1.80,0)};
\rowval{1}{+9.11}\rowval{2}{+4.25}\rowval{3}{+4.86}\rowval{4}{+3.50}\rowval{5}{+2.04}
\draw[black!60,line width=0.9pt] (axis cs:-5.7,6.2) -- (axis cs:-3.5,6.2);
\node[circle,fill=black!60,inner sep=0pt,minimum size=3.8pt] at (axis cs:-4.6,6.2) {};
\node[anchor=west,font=\scriptsize,inner sep=0pt] at (axis cs:-2.9,6.2) {estimate, $95\%$ interval};
\node[circle,draw=black!45,line width=0.35pt,inner sep=0pt,minimum size=2.6pt] at (axis cs:-4.6,7.0) {};
\node[anchor=west,font=\scriptsize,inner sep=0pt] at (axis cs:-2.9,7.0) {one seed};
\fill[black!12] (axis cs:4.4,6.78) rectangle (axis cs:6.6,7.22);
\node[anchor=west,font=\scriptsize,inner sep=0pt] at (axis cs:7.2,7.0) {$\pm2$ F1 margin};
\nextgroupplot[title={(b) Study II: induced roots, re-implementation},
  yticklabels={{total: real $-$ BPE},{granularity: scrambled $-$ BPE},
   {root: real $-$ scrambled},{real $-$ permuted},{real $-$ stratified},
   {permuted $-$ scrambled},{stratified $-$ scrambled}}]
\fill[black!12] (axis cs:-2,5.55) rectangle (axis cs:2,7.45);
\draw[black!45,line width=0.4pt] (axis cs:0,0.4) -- (axis cs:0,7.6);
\draw[black!35,densely dashed,line width=0.4pt] (axis cs:-6.5,3.5) -- (axis cs:18,3.5);
\addplot[seedpts] coordinates {
  (0.48,1) (-1.57,1) (-0.31,1) (0.06,1) (7.78,1)
  (-3.67,2) (-4.00,2) (-4.23,2) (-4.17,2) (-0.93,2)
  (4.15,3) (2.43,3) (3.92,3) (4.23,3) (8.71,3)
  (3.26,4) (-2.79,4) (3.72,4) (2.72,4) (8.09,4)
  (5.57,5) (-0.88,5) (4.10,5) (5.58,5) (10.97,5)
  (0.89,6) (5.22,6) (0.20,6) (1.51,6) (0.62,6)
  (-1.42,7) (3.31,7) (-0.18,7) (-1.35,7) (-2.26,7)};
\addplot[est=totcol] coordinates {(1.25,1) += (1.79,0) -= (1.72,0)};
\addplot[est=datcol] coordinates {(-3.39,2) += (1.89,0) -= (1.95,0)};
\addplot[est=tokcol] coordinates {(4.65,3) += (1.68,0) -= (1.64,0)};
\addplot[est=black!50] coordinates {(2.92,4) += (1.57,0) -= (1.60,0)
                                     (5.21,5) += (1.55,0) -= (1.56,0)
                                     (1.73,6) += (1.59,0) -= (1.54,0)
                                     (-0.56,7) += (1.55,0) -= (1.53,0)};
\rowval{1}{+1.25}\rowval{2}{-3.39}\rowval{3}{+4.65}\rowval{4}{+2.92}\rowval{5}{+5.21}
\rowval{6}{+1.73}\rowval{7}{-0.56}
\end{groupplot}
\end{tikzpicture}
\caption{Decomposition of the Amharic NER comparison with BPE (pooled readout, five seeds; in
Study~I, ``real'' is the analyser arm). The first three rows are the same contrasts in both studies:
the root component (blue) reproduces, whereas the granularity component (green) changes sign and
takes the total (black) with it. Rows below the dashed line are further controls of each study. The
far-right seed in (b) is one high-scoring \arm{real} run.}
\label{fig:root}
\end{figure*}

\subsection{Root assignment improves Amharic NER}
\label{sec:res-root}

In Study~I, the analyser tokenizer improves on BPE by $+9.11$ F1 under pooled readout
(Figure~\ref{fig:root}a). The scrambled control, which shares its vocabulary, fertility and
segmentation, reproduces $+4.25$ of this gain ($47\%$), and real root assignment adds the remaining
$+4.86$ $[3.14, 6.69]$ ($\pholm<10^{-4}$). Against the fertility-matched BPE control, which is not
vocabulary-matched, the root contribution is $+3.50$. The analyser-free induced arm reaches
$77.6\%$ of the gain of the analyser arm over BPE (F1 $23.81$ against $25.85$, from $16.74$).

Study~II reproduces the root effect against scrambling with an independently induced analyser:
$+4.65$ $[3.01, 6.33]$, positive in all five seeds ($\pseed=0.011$; Figure~\ref{fig:root}b). The
permutation controls reproduce the frequency profile of real roots as intended
(Appendix~\ref{app:study2}). Against the two permuted controls, real roots still score
higher over test sentences ($+2.92$ and $+5.21$, both $\pholm=0.006$). However, each of these
contrasts is negative in one seed ($\pseed=0.16$ and $0.055$), because one \arm{real} seed scores
$29.4$ against $21.3$--$23.7$ for the other four. Without this seed, real roots still beat
scrambled roots in all four remaining seeds (mean $+3.68$ F1, $\pseed=0.003$), whereas the
contrasts with the permuted and stratified controls fall to $+1.73$ and $+3.59$ F1 and are positive
in three of four seeds ($\pseed=0.34$ and $0.10$).

The part of the effect that frequency structure could explain is equivalent to zero at the
$2$-point margin for the stratified permutation ($-0.56$ $[-2.09, 0.99]$) and inconclusive for the
plain permutation ($+1.73$ $[0.19, 3.32]$, $\pholm=0.070$). The root effect against scrambling is
thus established in both studies and at the seed level. Frequency structure accounts for at most a small part of it, although the contrasts with the
permuted controls are sensitive to seed variation.

\subsection{Readout choice reverses comparisons}
\label{sec:res-readout}

\begin{figure*}[t]
\centering
\begin{tikzpicture}
\begin{axis}[
  paper, name=A, at={(0,0)}, anchor=south west,
  scale only axis, width=4.3cm, height=3.15cm,
  title={(a) Study I: difference from BPE under each readout}, title style={yshift=-4pt},
  xmin=-6.5, xmax=13.5, xtick={-5,0,5,10},
  ymin=0.45, ymax=4.55, y dir=reverse, ytick={1,...,4},
  yticklabels={{unigram ($15.4{\rightarrow}19.4$)},{induced ($12.2{\rightarrow}23.8$)},
               {lexicon ($14.3{\rightarrow}25.5$)},{analyser ($12.5{\rightarrow}25.9$)}},
  yticklabel style={font=\scriptsize}, ytick style={draw=none},
  xmajorgrids, ymajorgrids=false,
  xlabel={$\Delta$ entity F1 vs.\ BPE ($95\%$ interval)},
  legend style={at={(0.985,0.97)},anchor=north east},
]
\draw[black!45,line width=0.4pt] (axis cs:0,0.45) -- (axis cs:0,4.55);
\addplot[only marks,mark=*,mark size=1.9pt,color=tokcol,mark options={fill=white,draw=tokcol,line width=0.7pt},
  error bars/.cd,x dir=both,x explicit,error bar style={color=tokcol!60,line width=0.7pt},
  error mark options={rotate=90,mark size=1.4pt,line width=0.5pt}]
  coordinates {(0.70,0.84) += (1.64,0) -= (1.64,0)
               (-2.50,1.84) += (2.40,0) -= (2.28,0)
               (-0.47,2.84) += (2.54,0) -= (2.67,0)
               (-2.26,3.84) += (2.11,0) -= (2.06,0)};
\addlegendentry{first sub-token}
\addplot[est=tokcol]
  coordinates {(2.62,1.16) += (1.73,0) -= (1.60,0)
               (7.07,2.16) += (2.42,0) -= (2.41,0)
               (8.75,3.16) += (2.55,0) -= (2.48,0)
               (9.11,4.16) += (2.60,0) -= (2.53,0)};
\addlegendentry{mean pool}
\end{axis}
\begin{axis}[
  paper, name=B, at={(6.5cm,0)}, anchor=south west,
  scale only axis, width=6.3cm, height=3.15cm,
  title={(b) Readout gain by fertility and training regime}, title style={yshift=-4pt},
  xlabel={tokenizer fertility (tokens per word)},
  ylabel={readout gain $\Gamma$ (F1)}, ylabel style={yshift=-3pt},
  xmin=1.05, xmax=3.05, ymin=-6, ymax=20.5, ytick distance=5,
  grid=major,
  legend style={at={(0.015,0.97)},anchor=north west},
]
\draw[black!45,line width=0.4pt] (axis cs:1.05,0) -- (axis cs:3.05,0);
\addplot[only marks,mark=*,mark size=2pt,color=tokcol,mark options={fill=tokcol},
  error bars/.cd,y dir=both,y explicit,error bar style={tokcol,line width=0.6pt}]
  coordinates {(1.615,2.01) += (0,1.02) -= (0,0.99) (1.892,3.93) += (0,1.70) -= (0,1.69)
               (2.820,11.23) += (0,2.83) -= (0,2.76) (2.875,13.38) += (0,2.70) -= (0,2.64)
               (2.93,11.57) += (0,2.92) -= (0,2.95)};
\addlegendentry{Amharic, from scratch, Study I}
\addplot[only marks,mark=*,mark size=2pt,color=tokcol,mark options={fill=white,line width=0.7pt},
  error bars/.cd,y dir=both,y explicit,error bar style={tokcol,line width=0.6pt}]
  coordinates {(1.5345,4.28) += (0,1.62) -= (0,1.60) (2.10,4.68) += (0,1.68) -= (0,1.73)
               (2.13,2.73) += (0,1.56) -= (0,1.71) (2.16,4.23) += (0,1.65) -= (0,1.58)
               (2.19,3.09) += (0,1.60) -= (0,1.66)};
\addlegendentry{Amharic, from scratch, Study II}
\addplot[only marks,mark=square*,mark size=1.9pt,color=ctlcol,mark options={fill=ctlcol},
  error bars/.cd,y dir=both,y explicit,error bar style={ctlcol,line width=0.6pt}]
  coordinates {(1.187,-0.10) += (0,0.74) -= (0,0.77) (2.072,-2.98) += (0,1.84) -= (0,1.92)};
\addlegendentry{Hausa, from scratch, Study I}
\addplot[only marks,mark=diamond*,mark size=2.6pt,color=enccol,mark options={fill=enccol},
  error bars/.cd,y dir=both,y explicit,error bar style={enccol,line width=0.6pt}]
  coordinates {(2.25,0.54) += (0,1.56) -= (0,1.75) (2.30,0.72) += (0,2.23) -= (0,2.25)};
\addlegendentry{Amharic, pretrained, Study II}
\end{axis}
\end{tikzpicture}
\caption{The readout confound ($95\%$ intervals). (a) Under first-sub-token readout, no arm differs
significantly from BPE and every \rpt{} arm is nominally worse; under pooling, every arm is
significantly better ($\pholm\le0.006$). Row labels give entity F1 under the two readouts (BPE:
$14.7$ and $16.7$). (b) Gain from pooled over first-sub-token readout. The leftmost point of each
from-scratch series is BPE; points at nearly equal fertility are spread horizontally (Study~I
\rpt{} arms: $2.82$--$2.90$; Study~II \rpt{} arms: $2.17$; pretrained encoders: $2.20$, AfroXLM-R
left).}
\label{fig:readout}
\end{figure*}
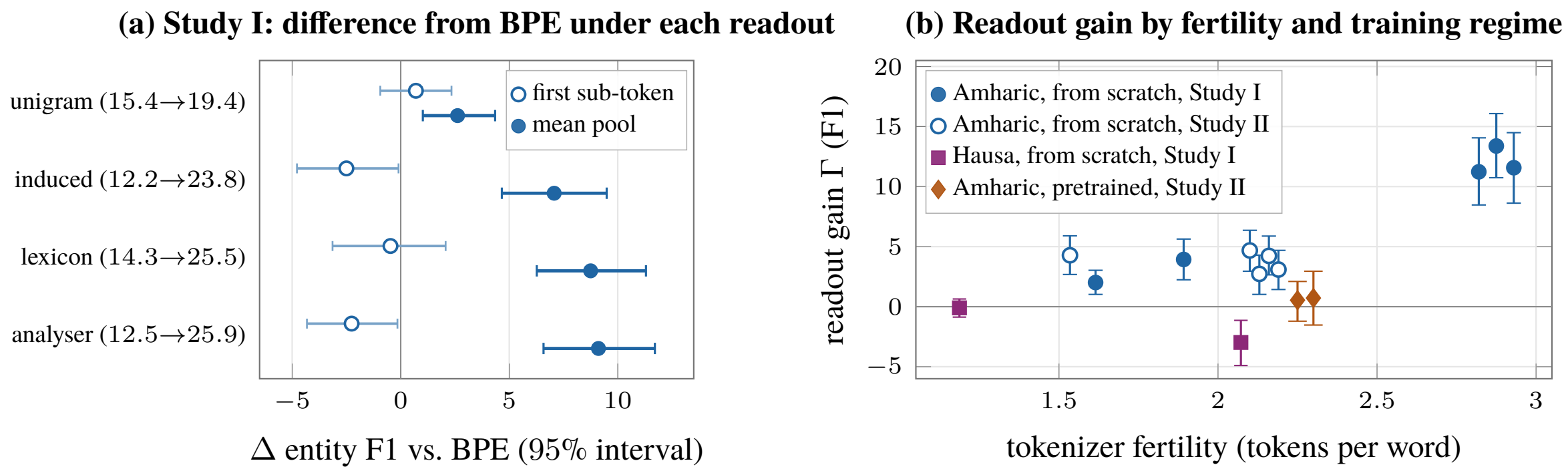

In Study~I, the \emph{direction} of every comparison between a \rpt{} arm and BPE depends on the
readout (Figure~\ref{fig:readout}a; Table~\ref{tab:readout1}). Under first-sub-token readout, every \rpt{} arm appears worse
than BPE ($-0.47$ to $-2.50$ F1, none significant after correction). Under pooling, every \rpt{} arm
is significantly better ($+7.07$ to $+9.11$). The reversal arises because pooling improves every
Amharic arm by a different amount: the readout gain grows from $+2.0$ F1 for BPE to $+13.4$ for the
analyser arm (Figure~\ref{fig:readout}b). By Equation~\ref{eq:readdecomp}, the readout
contamination of the comparison between the analyser arm and BPE is $+11.4$ F1, which is larger
than the comparison itself. Hausa shows the second regime described in Section~\ref{sec:readout}:
at BPE fertility $1.19$, the two readouts barely differ ($-0.10$, n.s.), and pooling hurts the
finer Hausa arms ($-2.98$, $\pholm=0.009$; $-4.11$ for unigram).

Study~II confirms that the readout gain is positive in every from-scratch arm ($+2.7$ to $+4.7$
F1, all intervals excluding zero). It does not confirm that the gain grows with fertility: across
a fertility contrast of $0.64$ tokens, the tokenizer-by-readout interactions against BPE are
undetectable ($+0.40$, $-1.56$, $-0.05$ and $-1.19$ F1; $\pholm\ge0.61$), and the trend across the
five arms of Study~I ($r=0.99$) should not be read as a scaling law. In the
two pretrained encoders, at fertility $2.20$, the readout gain is $+0.54$ $[-1.21, 2.10]$ for
AfroXLM-R, which is equivalent to zero at the $2$-point margin, and $+0.72$ $[-1.53, 2.95]$ for
XLM-R, which is inconclusive (Table~\ref{tab:pretrained}). Both gains are smaller than every
from-scratch gain at similar fertility, as expected for the contextualised regime of
Section~\ref{sec:readout}. The readout confound is therefore governed by the representation regime
rather than by fertility alone.

One XLM-R run diverged and was removed by the development-score gate; with it, the sentence-level
bootstrap would have reported a spurious readout effect of $-13.5$ F1 ($\pholm=0.001$;
Appendix~\ref{app:study2}).

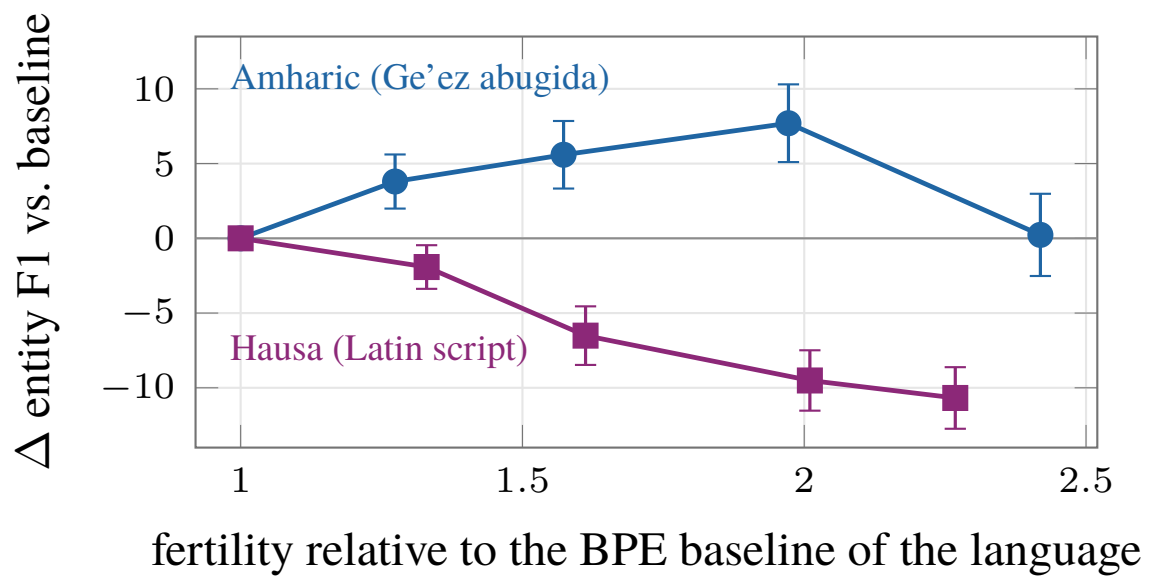
\begin{figure}[t]
\centering
\begin{tikzpicture}
\begin{axis}[
  paper, width=0.98\linewidth, height=4.3cm,
  xlabel={fertility relative to the BPE baseline of the language},
  ylabel={$\Delta$ entity F1 vs.\ baseline},
  xmin=0.92, xmax=2.52, ymin=-14, ymax=13.5, ytick distance=5,
  grid=major,
  every axis plot/.append style={line width=0.9pt,mark size=2pt},
]
\draw[black!45,line width=0.4pt] (axis cs:0.92,0) -- (axis cs:2.52,0);
\addplot+[tokcol,mark=*,mark options={fill=tokcol},
  error bars/.cd,y dir=both,y explicit,error bar style={line width=0.6pt}]
  coordinates {(1.000,0.00) +- (0,0) (1.274,3.80) +- (0,1.81) (1.573,5.59) +- (0,2.26)
               (1.972,7.70) +- (0,2.60) (2.419,0.23) +- (0,2.75)};
\addplot+[ctlcol,mark=square*,mark options={fill=ctlcol},
  error bars/.cd,y dir=both,y explicit,error bar style={line width=0.6pt}]
  coordinates {(1.000,0.00) +- (0,0) (1.330,-1.92) +- (0,1.46) (1.612,-6.51) +- (0,1.96)
               (2.010,-9.51) +- (0,2.02) (2.268,-10.68) +- (0,2.06)};
\node[anchor=west,font=\scriptsize,text=tokcol,inner sep=1pt] at (axis cs:0.97,10.6)
  {Amharic (Ge'ez abugida)};
\node[anchor=west,font=\scriptsize,text=ctlcol,inner sep=1pt] at (axis cs:0.97,-7.6)
  {Hausa (Latin script)};
\end{axis}
\end{tikzpicture}
\caption{Granularity alone: NER F1 of capacity-matched BPE tokenizers relative to the baseline of
each language (Study~I, five seeds, $95\%$ intervals). Every non-baseline point except Amharic at
$2.42\times$ is significant after Holm correction.}
\label{fig:gran}
\end{figure}

\subsection{Granularity changes sign}
\label{sec:res-gran}

At matched capacity, finer BPE segmentation has opposite effects on NER in the two languages
(Figure~\ref{fig:gran}). Amharic improves up to $1.97$ times its baseline fertility ($+7.7$ F1) and
returns to the baseline at $2.42$ times ($+0.2$, $\pholm=0.85$). Hausa declines at every level, to
$-10.7$ F1. These results are consistent with an orthographic explanation, although two languages
cannot establish it: finer units help when the script fuses consonant and vowel, and they hurt when
the default units are already close to whole words, as in Hausa ($1.19$ BPE tokens per word).

The granularity step of the \rpt{} tokenizer is less stable still. Scrambled minus BPE is $+4.25$
F1 in Study~I but $-3.39$ in Study~II, in which the scrambled arm has lower fertility ($2.17$
against $2.89$), a different analyser and lower coverage. Both fertility ratios lie on the rising
part of the Study~I BPE sweep, and we have not identified the cause of the reversal. Its
consequence is the problem this paper addresses: in Study~II, the total comparison of real roots
with BPE is $+1.25$ F1 and inconclusive ($\pholm=0.15$), and a two-condition study would have
reported no benefit. The decomposition shows instead that a root gain and a granularity loss of
similar size cancel (Figure~\ref{fig:root}b), which is the situation described by
Equation~\ref{eq:decomp}.

\subsection{Translation}
\label{sec:res-mt}

In the Study~I translation systems ($16.9$ and $31.1$ \chrf{} with BPE for Amharic and Hausa),
finer target segmentation hurts in both languages. The Amharic \rpt{} arms lose $0.95$--$1.47$
\chrf{}, the scrambled control loses $0.90$, and the Hausa fertility-matched BPE control loses
$1.52$ (all $\pholm<10^{-3}$; Appendix~\ref{app:study1}). Real and scrambled roots do not differ
($-0.05$ \chrf{}, $\pholm=0.30$), but the analyser covers only $59.7\%$ of MT types, and a root
contrast is attenuated in proportion to coverage (Appendix~\ref{app:mt}). We therefore draw no
conclusion about root assignment in translation: the root and readout findings apply to token-level
classification, and for generation the only result we consider established is the cost of finer
target segmentation.

\begin{table*}[t]
\centering
\small
\setlength{\tabcolsep}{3.5pt}
\begin{tabular}{llll}
\toprule
\textbf{Study I conclusion} & \textbf{Study I} & \textbf{Study II} & \textbf{Verdict} \\
\midrule
Real roots beat scrambled roots (NER) & $+4.86$ F1 & $+4.65$ F1, positive in $5/5$ seeds & reproduced \\
Readout gain positive in every arm & $+2.0$ to $+13.4$ F1 & $+2.7$ to $+4.7$ F1; pretrained $<1$ & reproduced in sign \\
Readout gain scales with fertility & $r=0.99$ (5 arms) & interactions undetectable & not reproduced \\
\rpt{} granularity step helps & $+4.25$ F1 & $-3.39$ F1 & not reproduced \\
Encoder channel helps behind BPE & $+1.83$ F1 & $-0.42$ F1, equivalent to zero & not reproduced \\
No root effect on MT & $-0.05$ \chrf & $+0.19$ \chrf{} (sentences), $\pseed=0.11$ & null not confirmed \\
\bottomrule
\end{tabular}
\caption{Replication ledger. ``Reproduced'' requires the same sign, a comparable size relative to
the between-seed variability of each study, and a consistent verdict. The encoder-channel analysis
is described in Appendix~\ref{app:channel}.}
\label{tab:repl}
\end{table*}

\subsection{Replication ledger}
\label{sec:res-repl}

Table~\ref{tab:repl} lists the Study~I conclusions that Study~II tested: two held, three did not,
and the MT null was neither confirmed nor overturned. Four further predictions failed within
Study~I: the transfer of the granularity gain to Hausa, a readout effect that holds for every
tokenizer, a compression advantage of the \rpt{} tokenizer, and a benefit of root-based attention
biases for NER. Nine nulls are equivalences at the pre-specified margins
(Appendix~\ref{app:study2}), and all other nulls are only inconclusive.

\subsection{Discussion and recommendations}
\label{sec:disc}

\paragraph{The root effect concerns lexical identity.} The scrambled control changes only which
word forms share a root token. The $+4.9$ and $+4.7$ F1 gains therefore measure the value of
grouping the forms of a root, on a task that is not itself morphological. A plausible account,
which we did not test, is that a root token pools evidence across forms that $1{,}750$ training
sentences attest only sparsely.

\paragraph{Scope of the readout confound.} A controlled tokenizer comparison has to train a model
with each tokenizer, and new tokenizers are evaluated in the same way. Our largest effect thus
appears in the regime in which tokenizer comparisons are made. The pretrained result bounds this
finding rather than diluting it: a practitioner who fine-tunes XLM-R or AfroXLM-R on Amharic NER can
treat the readout as a minor choice, whereas a study that trains models from scratch cannot.

\paragraph{The components vary with the configuration.} The two implementations agree on the root
effect and on the presence of a readout confound, and they disagree on the granularity step and on
the fertility trend of the readout gain. The components of a tokenizer effect therefore have to be
measured in the configuration of interest rather than inferred from a total.

\paragraph{Recommendations.} Tokenizer comparisons on token-level tasks should cross the readout
with the tokenizer and match vocabulary, non-embedding parameters and word-level sequence budgets. A claim about morphological content should be supported by a
scrambled-root control, preferably with a frequency-preserving permutation, and by a
capacity-matched granularity sweep. The sweep also gives a practical rule for vocabulary size: the
effect of granularity should be measured in the target language before a size is chosen, since
finer segmentation helped Amharic NER up to about $0.8$ BPE tokens per fidel and hurt Hausa NER at
every level. Finally, studies should report per-seed scores with an equivalence verdict for every
null, ideally from a two-level bootstrap over training seeds and test sentences. For low-resource
practice, about $78\%$ of
the analyser-backed NER benefit was reachable from a paradigm list alone, and paradigm lists exist
for far more languages than analysers do.
\section{Conclusion}
\label{sec:concl}

A morphology-aware tokenizer changes segmentation granularity, root grouping and, on token-level
tasks, the conditions under which word labels are read. With controls that change one factor at a
time, in two independent implementations, we found that real root assignment improves
Amharic NER beyond scrambled roots, with little of the gain due to frequency structure. Readout
choice reverses the sign of every comparison between the root-and-pattern tokenizers and BPE in
from-scratch models, whereas it matters little in two pretrained encoders. The granularity
component changes sign across languages and implementations. Tokenizer effects should therefore be
decomposed before a gain is attributed to linguistic structure, with the readout and the
representation regime reported. Whether root factorisation remains beneficial under large-scale
pretraining is an open question.
\section*{Limitations}
\label{sec:limitations}

First, the root result rests on one language, measured with two analysers. A Tigrinya extension
could not be run for lack of an NER corpus. The root finding should therefore be read as evidence
for Amharic, not for root-and-pattern languages in general, and the orthographic explanation of the
granularity effect rests on two languages only.

Second, the root and granularity effects were measured with small encoders (about $1$M to $2$M
parameters for NER) trained from scratch on $1{,}750$ sentences, and absolute F1 is correspondingly
low. This is the setting in which the controls can be run, but the effects have not been tested
with larger models or with continued pretraining. The pretrained experiments cover the readout and the encoder
channel only, in two encoders with three seeds.

Third, with five seeds, a single run can dominate a contrast. The contrasts between real roots and
the permutation controls are significant over test sentences but not over seeds, and we did not run
a two-level bootstrap that resamples seeds and sentences jointly.

Fourth, our evidence on translation is limited to the granularity effect. The MT corpora contain
about $44\%$ religious text, decoding is greedy, and the Study~II systems are near the floor for
reasons we have not diagnosed. The NER findings, which use news text, do not depend on these
limitations, but they are restricted to one token-level task and one dataset per language.

Fifth, two parts of the design are incomplete. The cause of the sign reversal of the granularity
step between the two studies was not identified, and BPE learned over the decomposed
(consonant, vowel) stream, a natural intermediate baseline, remains untested.

\section*{Ethical considerations}

This work aims to improve language technology for languages that current pipelines serve less well.
Its most practical finding is that much of the benefit of morphological tokenization is reachable
from paradigm lists rather than analysers, which lowers the resource barrier for such languages.
Results on two languages should not be generalised to ``African'' or ``Afroasiatic'' languages as a
category. The translation data are dominated by religious and software-localisation registers, and
any system built on them would reflect those registers. NER can support surveillance or profiling
when applied to personal communications; our models are small research instruments trained on
public news data, and they are not released as tools. The MasakhaNER data were created by the
Masakhane community, whose request to credit contributing annotators we follow. HornMorpho is
GPL-licensed and is used only as an external data source behind a process boundary. We report the
divergent fine-tuning run rather than discarding it. Total logged training time was about $10$
hours on a single commodity GPU.

\bibliography{references}

\appendix
\section{Properties of the decomposition and the controls}
\label{app:proofs}

Let $S_a(X)$ be the expected test score with tokenizer $X$ and readout $a$. The components $\Dg$
and $\Dr$ are defined by Equation~\ref{eq:decomp} for a reference tokenizer $\Tok^{\circ}$.

\begin{proposition}[Two conditions do not identify the root component]
\label{prop:nonid}
The observations $(S_a(B),S_a(\Tok))$ are consistent with every value of $\Dr$, because setting
$S_a(\Tok^{\circ}):=S_a(\Tok)-c$ leaves both observed scores unchanged for any $c\in\mathbb{R}$.
\end{proposition}

\paragraph{Controls.} Let $\rho$ assign each analysable word type $w\in\mathcal{W}_\Rt$ its root
identifier in $\mathcal{V}_\Rt$. The scrambled control draws
$\tilde\rho(w)\sim\mathrm{Unif}(\mathcal{V}_\Rt)$ independently for each type. The permuted control
uses $\rho_\pi=\rho\circ\pi$ for a uniformly random permutation $\pi$ of $\mathcal{W}_\Rt$. The
stratified variant draws an independent permutation within each decile of corpus frequency. All
controls replace only the root token.

\begin{proposition}[What the controls hold fixed]
\label{prop:scramble}
(i)~Both controls use the token inventory of $\Tok$ and emit the same number of tokens for every
word, hence the same fertility, parameter count and token positions. Their root assignment is
independent of morphology. (ii)~Under scrambling, the number of types per root identifier matches
the real assignment only in mean and does not reproduce the skew of real roots. (iii)~Under
permutation, this number equals the real count exactly for every draw. Token-weighted identifier
frequencies are preserved in expectation, within each frequency decile for the stratified variant.
\end{proposition}
\begin{proof}
(i)~The substitution draws identifiers from $\mathcal{V}_\Rt$ and rewrites exactly the positions
that hold root tokens, one token for one. No inventory entry is therefore added, and every token
sequence keeps its length. Fertility, positions and the embedding table are unchanged. The
scrambled draw depends only on the seed. For a fixed $w$, the permuted identifier $\rho(\pi(w))$ is
distributed as $\rho$ at a uniformly random type. Neither assignment therefore depends on
morphology. (ii)~Each type picks an identifier uniformly and independently. The number of types per
identifier is thus binomial with $|\mathcal{W}_\Rt|$ trials and success probability
$1/|\mathcal{V}_\Rt|$. Its mean $|\mathcal{W}_\Rt|/|\mathcal{V}_\Rt|$ equals the mean under $\rho$,
but the realised counts are approximately uniform rather than skewed. (iii)~For an identifier $r$,
$\{w:\rho(\pi(w))=r\}=\pi^{-1}(\{w':\rho(w')=r\})$. A bijection preserves cardinality, and the count
therefore equals the real count $n_r$ for every $\pi$. With $c(w)$ the corpus frequency and
$\bar c$ its mean,
\begin{align*}
\mathbb{E}\Big[\sum_w c(w)\,\mathbf{1}[\rho(\pi(w))=r]\Big]
&=\sum_w c(w)\,\frac{n_r}{|\mathcal{W}_\Rt|}\\
&=n_r\bar c .
\end{align*}
Under the stratified variant, the same computation holds within each decile block, and the
within-block dispersion of $c$ is at most the global one.
\end{proof}

\paragraph{Readout.} For a word split into $m$ sub-tokens, let $h_i=\mu_w+\beta_i+\varepsilon_i$,
with $\mu_w$ the representation the classifier needs, $\beta_i$ a fixed fragment offset and
$\varepsilon_i$ noise with $\mathbb{E}\varepsilon_i=0$, variance $\sigma^2$ and pairwise
correlation $\varrho$. First-sub-token readout uses $h_1$, whereas pooled readout uses
$\bar h=\frac1m\sum_ih_i$. We write $\bar\beta=\frac1m\sum_i\beta_i$.

\begin{proposition}[Readout bias and variance]
\label{thm:readout}
$\mathrm{MSE}_{\mathrm{first}}-\mathrm{MSE}_{\mathrm{pool}}$ with respect to $\mu_w$ equals
Expression~\ref{eq:readout}. Its absolute value is at most
$|\beta_1^2-\bar\beta^2|+\sigma^2(1-\varrho)$, which tends to zero when the offsets equalise and
$\varrho\to1$, for any $m$.
\end{proposition}
\begin{proof}
$h_1-\mu_w=\beta_1+\varepsilon_1$ has second moment $\beta_1^2+\sigma^2$. Further,
$\bar h-\mu_w=\bar\beta+\bar\varepsilon$ with
\begin{align*}
\mathrm{Var}(\bar\varepsilon)&=\tfrac{1}{m^2}\big(m\sigma^2+m(m-1)\varrho\sigma^2\big)\\
&=\tfrac{\sigma^2}{m}\big(1+(m-1)\varrho\big).
\end{align*}
Subtracting, and using
$\sigma^2-\frac{\sigma^2}{m}(1+(m-1)\varrho)=\sigma^2(1-\varrho)\frac{m-1}{m}$, gives
Expression~\ref{eq:readout}. The bound follows from $\frac{m-1}{m}\le1$. For $m=1$, the readouts
coincide. Every multi-token word has at least two tokens, and the fraction of split word tokens is
therefore at most the fertility minus one, which is $19\%$ for Hausa BPE.
\end{proof}

\section{The root-and-pattern tokenizer}
\label{app:rootpat}

\paragraph{Script decomposition.} In the core Ethiopic block of Unicode, the fidel of one consonant
occupy eight consecutive code points in the vowel order \"a, u, i, a, e, \schwa, o,
$^{\mathrm{w}}$a \citep{unicode15}, which gives the decomposition of Section~\ref{sec:tok}. We
verified this layout on all $231$ fidel attested in UniMorph Amharic. The script map $\delta$
replaces each Ethiopic character by its pair $(\row,\ord)$ and carries any other character as a
literal. Its inverse $\gamma$ recomposes the string. The composition $\gamma\circ\delta$ is the
identity on all $53{,}031$ attested UniMorph and MasakhaNER types. For a word $w$, $\cons(w)$ is
its sequence of rows. Amharic orthography does not mark gemination \citep{leslau1995amharic}, and
each radical realised in a word is consequently written by exactly one fidel.

\begin{theorem}[Sub-character opacity]
\label{thm:opacity}
Let $w$ and $w'$ realise a root $\Rt=(r_1,\dots,r_k)$ with radicals at fidel positions
$p_1<\dots<p_k$ in $w$ and $p'_1<\dots<p'_k$ in $w'$. If $\ord(w_{p_j})\neq\ord(w'_{p'_j})$ for
some $j$, no character string containing radical $j$ occurs at the position of that radical in
both forms. In particular, no vocabulary of character substrings contains a unit that spells the
root in both forms.
\end{theorem}
\begin{proof}
A fidel fixes both its row and its order. Any string containing radical $j$ at the position of the
radical must contain $w_{p_j}$ when it occurs in $w$ and $w'_{p'_j}$ when it occurs in $w'$. These
fidel share a row but differ in order, and they are therefore different characters. A unit spelling
the whole root contains every radical, including radical $j$.
\end{proof}

\paragraph{Templates and ghost radicals.} The template $\hat\pi$ records, slot by slot across the
stem, a radical with its vowel order $(\textsf{R},o)$, a non-root fidel with its row and order
$(\textsf{U},r,o)$, a literal $(\textsf{L},c)$, or a \emph{ghost} radical $(\textsf{G})$ that is
consumed from the root without a surface exponent, such as the medial glide of \gz{ሞተ}{mot\"a}
`die'.

\begin{theorem}[Losslessness]
\label{thm:lossless}
Let $\nu$ be homophone normalisation. For every analysable word $w$, interleaving the root and
template produced by \textup{\textsc{Analyse}}, with ghost slots consuming a radical and emitting
nothing, returns $\nu(w)$ exactly. The fallback path is also lossless. Hence \rpt{} is lossless
modulo $\nu$.
\end{theorem}
\begin{proof}
Write $\delta(\nu(w))=u_1\dots u_n$ and let $[i_f,i_l]$ be the stem span located by
\textup{\textsc{Analyse}}. Prefix and suffix are stored verbatim. Within the span, the algorithm
emits one slot per unit in order. In each case, the pair $(\row,\ord)$ is recoverable, with the row
of a radical taken from the root. The map $\gamma$ therefore reconstructs the unit through
$u=\texttt{0x1200}+8\row+\ord$. A ghost slot consumes one radical without emitting a unit. The
inverse thus consumes the root iterator exactly $k$ times while emitting $k-g$ radicals for $g$
ghosts. Ghosts are placed immediately before the next realised radical and do not perturb the order
of emitted units. Concatenation gives $\delta(\nu(w))$, from which $\gamma$ returns $\nu(w)$. The
fallback is a segmentation over a closed unit alphabet with byte fallback, which concatenates back
to its input.
\end{proof}

\noindent
Normalisation merges homophone rows that modern Amharic pronounces identically
\citep{belay2022normalization}, affecting $5.9\%$ of types.

\paragraph{Analysis procedure.}\label{alg:analyse} \textsc{Analyse} first maps a word $w$ to its
unit sequence $U=\nu(\delta(w))$ and splits $U$ into its rows (the consonantal tier) and vowel
orders. It then searches the consonantal tier for a root from the inventory under the gap
constraint described below, considering repaired candidates only when no unrepaired analysis
exists. A radical that falls on a repair position becomes a ghost. The units that precede the first
realised radical are stored as the prefix, and those that follow the last one as the suffix. Every unit of the stem span
between them then receives one template slot, and each ghost slot is placed immediately before the
next realised radical. A word without a root match is passed to the fallback segmentation.

\paragraph{Root search.} \textsc{Search} accepts a root $\hat\Rt$ at positions $p_1<\dots<p_k$ of
the consonantal tier only if $p_{j+1}-p_j\le\kappa+1$ for every $j$. Candidates are scored by
$2|\hat\Rt|-(\mathrm{span}-|\hat\Rt|)-1.5\,\mathrm{cost}-0.5\cdot\mathbf{1}[|\hat\Rt|=2]$. Without
the per-gap bound, a triliteral root could match an arbitrary word by scattering its radicals; a
unit test asserts that such a match is rejected. A repaired candidate is considered only when no
unrepaired analysis exists. An early version lacked this rule and attributed $91\%$ of its
supposedly gemination-sensitive analyses to one repair heuristic.

\paragraph{Root sources.} The induced arm keeps roots of three to five radicals, obtained as the
longest common subsequence (LCS) of the consonantal tiers of a paradigm, from $2{,}259$ of
$3{,}041$ training paradigms; the other $760$ paradigms are held out before induction. Of the
induced roots, $59\%$ have four or more radicals because induction absorbs paradigm-constant
prefixes. The lexicon arm consults the $11{,}529$ noun stems of HornMorpho before its verb roots.
It disables ghost radicals, because allowing one ghost raised verb coverage from $65.4\%$ to
$97.6\%$ while leaving $92.1\%$ of ghosted matches with only two real radicals; held-out paradigm
cohesion also fell with ghosts. The analyser arm selects the root by majority vote, because
$38.0\%$ of types have several analyses, of which only $66.5\%$ agree on the root. This arm allows
one lexicon-licensed ghost radical as well as the Amharic palatalisation alternations
\gz{ድ}{d}$\sim$\gz{ጅ}{\v{g}}, \gz{ት}{t}$\sim$\gz{ች}{\v{c}}, \gz{ስ}{s}$\sim$\gz{ሽ}{\v{s}},
\gz{ዝ}{z}$\sim$\gz{ዥ}{\v{z}}, \gz{ን}{n}$\sim$\gz{ኝ}{\~{n}} and \gz{ል}{l}$\sim$\gz{ይ}{y}. Before
lookup, the non-standard \gz{ዉ}{w} (\texttt{U+12C9}) is mapped to \gz{ው}{w} (\texttt{U+12CD}),
which recovers $285$ types. The analyser arm routes $73.6\%$ of NER types and $71.1\%$ of tokens
through the factored path. Held-out paradigm cohesion is $0.885$ (induced), $0.924$ (lexicon) and
$0.977$ (analyser) overall, and $0.673$, $0.763$ and $0.962$ for verbs.

\section{Secondary analysis: encoder-side linguistic features}
\label{app:channel}

\paragraph{Background.} Factored models add embedding channels for linguistic annotations
\citep{sennrich2016factors}. \citet{armengol2021factors} report gains of $0.8$--$1.2$ BLEU from
such factors in a Transformer, whereas \citet{gessler2023syntactic} find that syntactic inductive
biases give little benefit in most low-resource settings. Related work on the input representation
pursues a similar principle by other means: \citet{samuel2023factorizer} factorise each sub-word
into a triplet of discrete codes, and \citet{kumar2026bytes} combine byte-level chunking with
lightweight part-of-speech supervision. At the sequence level, \citet{ekanayake2026helabert}
propose a dual pooling classification head for Sinhala. Together, these studies motivate evaluating
a linguistic channel by the information it adds beyond the token stream.

An encoder-side embedding channel is often argued to help only if its feature cannot be derived
from the surface string. Amharic gemination is such a feature: \gzt{ይሰማሉ}{y\schwa s\"amallu}
receives two analyses, `they hear' and `they are heard', that differ only in gemination. We tested
a gated channel $h_i\leftarrow h_i+\tanh(g)(E_a(f_a(i))+E_b(f_b(i)))$, with $g$ initialised at
zero, in two parameter-identical versions ($1{,}665$ parameters each). The \emph{non-derivable}
channel carries three-valued gemination and root class from the analyser. The \emph{derivable}
channel carries the consonant--vowel template class, which the string determines. In Study~II,
whose analyser has no gemination information, the non-derivable feature $F$ is a root-class
stand-in with $8$ classes, and the derivable control $D$ is the row of the final fidel modulo $10$.

\paragraph{A bound.} Let $W$ be a word occurrence, $Y$ its label and $F=f(W)$ a feature.

\begin{theorem}[Redundancy bound]
\label{thm:redbound}
Let $\mathcal{G}(f\mid\Tok)=H(Y\mid\Tok(W))-H(Y\mid\Tok(W),F)$ be the reduction in Bayes-optimal
log loss from adding $F$. Then
\begin{align*}
0\le\mathcal{G}(f\mid\Tok)&=I(Y;F\mid\Tok(W))\\
&\le H(F\mid\Tok(W)).
\end{align*}
If $\Tok$ is lossless and $f$ is a function of the surface string, $H(F\mid\Tok(W))=0$.
\end{theorem}
\begin{proof}
The equality is the definition of conditional mutual information, which is non-negative. The upper
bound follows from
$I(Y;F\mid\Tok(W))=H(F\mid\Tok(W))-H(F\mid Y,\Tok(W))\le H(F\mid\Tok(W))$
\citep{cover2006elements}. If $\Tok$ is lossless, the surface string is a function of $\Tok(W)$,
as is any function of that string.
\end{proof}

\noindent
An analyser-backed tokenizer can deliver information the string lacks, and the bound can therefore
be small even for a non-derivable feature. Ordering tokenizers by the quality of root information
they supply (BPE and unigram $\prec$ induced $\prec$ lexicon $\prec$ analyser), the string-based
account predicts a constant non-derivable channel effect $\Dnd$, whereas the redundancy account
predicts that $\Dnd$ decreases along this order. A finite model with sentence context need not
realise a word-level Bayes bound; the theorem implies only that $\Dnd$ should be negligible
wherever $H(F\mid\Tok(W))$ is small.

\paragraph{Estimating the bound.} The plug-in estimate of $H(F\mid\Tok(W))$ is near zero for every
lossless tokenizer ($\le0.005$ bits in all arms), which makes it uninformative. We instead fit a
multinomial logistic model $q(F\mid\text{bag of tokens})$ on $K=5$ folds of the training vocabulary
and report the frequency-weighted held-out log loss $\Hcv$.

\begin{lemma}[$\Hcv$ estimates an upper bound]
\label{lem:hcv}
For any function $Z$ of $\Tok(W)$ and any $q$,
$\mathbb{E}[-\log_2 q(F\mid Z)]\ge H(F\mid Z)\ge H(F\mid\Tok(W))$.
\end{lemma}
\begin{proof}
The expected log loss decomposes as
$H(F\mid Z)+\mathbb{E}[D_{\mathrm{KL}}(p(\cdot\mid Z)\,\|\,q(\cdot\mid Z))]$,
and conditioning on a refinement does not increase entropy \citep{cover2006elements}.
\end{proof}

\begin{proposition}[Root-randomising controls are not invertible]
\label{prop:ctlred}
Suppose that two analysable types share prefix, template and suffix and receive the same root
identifier from a control. The control then maps them to the same token sequence. As a result,
$H(f(W)\mid\Tok^{\circ}(W))$ can be positive for a string-derivable $f$, although
$H(f(W)\mid\Tok(W))=0$.
\end{proposition}
\begin{proof}
The token sequence is determined by prefix, root identifier, template and suffix. If all four agree
for $w\neq w'$, the map is not injective, and $f$ may take several values on such a fibre. Under
the real assignment, two distinct types cannot share all four (Theorem~\ref{thm:lossless}).
\end{proof}

\noindent
The controls therefore remain exact references for the root contrast of
Section~\ref{sec:res-root}, which does not require invertibility. They are not neutral references
for a channel experiment.

\begin{figure}[t]
\centering
\begin{tikzpicture}
\begin{axis}[
  paper, width=\linewidth, height=5.0cm,
  xlabel={$\Hcv(F\mid\Tok)$ (bits)},
  ylabel={channel effect $\Dnd$ (F1)},
  xmin=0.1, xmax=2.0, ymin=-2.4, ymax=7.5, grid=major,
  tick label style={font=\scriptsize}, label style={font=\footnotesize},
  legend style={font=\scriptsize,at={(0.02,0.97)},anchor=north west,legend columns=3,
                inner sep=1.5pt}, legend cell align=left]
\fill[black!12] (axis cs:0.1,-2) rectangle (axis cs:2.0,2);
\draw[black!45,line width=0.4pt] (axis cs:0.1,0) -- (axis cs:2.0,0);
\addplot+[only marks,tokcol,mark=*,mark size=2.4pt,mark options={fill=tokcol},
  error bars/.cd, y dir=both, y explicit, error bar style={tokcol}]
  coordinates {(0.320,0.59) += (0,1.43) -= (0,1.44)};
\addlegendentry{real roots}
\addplot+[only marks,ctlcol,mark=square*,mark size=2.2pt,mark options={fill=ctlcol},
  error bars/.cd, y dir=both, y explicit, error bar style={ctlcol}]
  coordinates {(1.084,2.30) += (0,1.71) -= (0,1.73) (1.020,1.87) += (0,1.61) -= (0,1.50)
               (1.15,2.75) += (0,1.53) -= (0,1.38)};
\addlegendentry{randomised roots}
\addplot+[only marks,enccol,mark=triangle*,mark size=2.9pt,mark options={fill=enccol},
  error bars/.cd, y dir=both, y explicit, error bar style={enccol}]
  coordinates {(1.795,-0.42) += (0,1.38) -= (0,1.40)};
\addlegendentry{BPE}
\end{axis}
\end{tikzpicture}
\caption{Study~II: feature entropy left undetermined by the tokens (held-out estimate $\Hcv$)
against the measured effect of the non-derivable channel (pooled readout, $95\%$ intervals). The
three randomised-root arms ($\Hcv$ between $1.06$ and $1.08$) are spread horizontally for
legibility. Shaded band: $\pm2$ F1 equivalence margin. The bound forbids large effects at small
$\Hcv$ only.}
\label{fig:channel}
\end{figure}
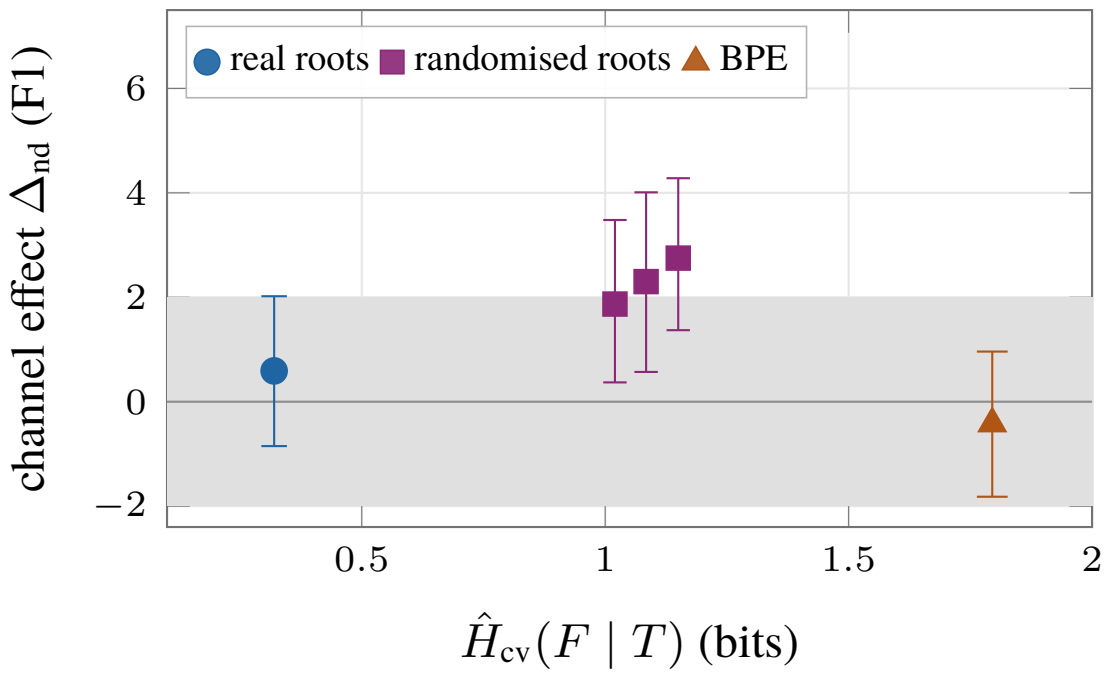

\paragraph{Results.} In Study~I, the derivable channel has no detectable effect behind any
tokenizer. The non-derivable channel improves NER behind the two tokenizers that supply no root
information (BPE $+1.83$, unigram $+2.26$, both significant). Its effect is smaller behind the
induced arm ($+1.30$, $\pholm=0.18$), and it has no detectable effect behind the lexicon and
analyser arms. This pattern is inconsistent with a constant effect and consistent with the
redundancy account, although the gradient is not strictly monotone. Three attention biases linking
sub-tokens with the same root, root class or radical had no detectable effect
($|\Delta|\le0.12$ F1), although each raised attention mass on its target pairs by
$1.65$--$2.44\times$.

In Study~II (Figure~\ref{fig:channel}, Table~\ref{tab:chan2}), the tokens leave $\Hcv=1.80$ of the
$2.17$ bits of the feature undetermined behind BPE, $0.32$ bits behind real roots and
$1.06$--$1.08$ bits behind the randomised arms, consistent with Proposition~\ref{prop:ctlred}.
Behind real roots, for which the bound is tight, the channel effect is $+0.59$ F1, which is
equivalent to zero. Behind the scrambled and stratified arms, the channel helps ($+2.30$,
$+2.75$). Behind BPE, for which the bound is largest, the effect is $-0.42$, which is also
equivalent to zero. The ordering prediction, which Study~I had supported, therefore fails. The
bound thus indicates when a channel cannot help and is silent on when it will. The derivable
channel also helps behind two non-invertible controls ($+3.30$ and $+2.68$), as
Proposition~\ref{prop:ctlred} permits. However, it reaches $+1.64$ (n.s.) behind BPE as well, and
the data consequently do not single out the redundancy account. In both pretrained encoders,
neither channel helps (Table~\ref{tab:pretrained}).

\begin{table*}[t]
\centering
\footnotesize
\setlength{\tabcolsep}{3pt}
\resizebox{\textwidth}{!}{%
\begin{tabular}{lccccc}
\toprule
& & \multicolumn{2}{c}{\textbf{Non-derivable $F$}} & \multicolumn{2}{c}{\textbf{Derivable $D$}} \\
\cmidrule(lr){3-4}\cmidrule(lr){5-6}
\textbf{Arm} & $\Hcv$ (bits) & \textbf{main}, $\pholm$ & \textbf{interaction} & \textbf{main}, $\pholm$ & \textbf{interaction} \\
\midrule
\arm{bpe}       & $1.795$ $[1.696,1.890]$ & $-0.42$ $[-1.82,+0.96]$, $0.810^{\mathrm{eq}}$ & -- & $+1.64$ $[+0.04,+3.15]$, $0.192$ & -- \\
\arm{scrambled} & $1.084$ $[0.989,1.188]$ & $+2.30$ $[+0.57,+4.01]$, $0.035$ & $+2.72$ $[+0.54,+4.94]$ & $+3.30$ $[+1.75,+4.99]$, $0.010$ & $+1.66$ $[-0.43,+3.91]$ \\
\arm{perm}      & $1.056$ $[0.955,1.161]$ & $+1.87$ $[+0.37,+3.48]$, $0.168$ & $+2.29$ $[+0.39,+4.21]$ & $+1.65$ $[+0.13,+3.14]$, $0.168$ & $+0.00$ $[-2.04,+2.09]^{\mathrm{eq}}$ \\
\arm{permstrat} & $1.062$ $[0.963,1.167]$ & $+2.75$ $[+1.37,+4.28]$, $0.010$ & $+3.17$ $[+1.35,+5.03]$ & $+2.68$ $[+1.26,+4.11]$, $0.010$ & $+1.03$ $[-0.81,+3.12]$ \\
\arm{real}      & $0.320$ $[0.307,0.334]$ & $+0.59$ $[-0.85,+2.02]$, $0.810^{\mathrm{eq}}$ & $+1.01$ $[-0.91,+2.91]$ & $+1.18$ $[-0.26,+2.72]$, $0.327$ & $-0.46$ $[-2.40,+1.52]$ \\
\bottomrule
\end{tabular}}
\caption{Study~II redundancy estimates and channel effects (pooled readout, F1 $\times100$). $\Hcv$
uses $K=5$ folds and $500$ bootstrap resamples; the marginal entropy of $F$ is $2.165$ bits. The
plug-in estimate is at most $0.005$ bits in every arm. Interactions are double differences against
\arm{bpe}; Holm correction within families of five (main) and four (interaction).
$^{\mathrm{eq}}$ equivalent to zero at $\delta=2$; no contrast is equivalent at $\delta=1$.}
\label{tab:chan2}
\end{table*}

\section{Secondary analysis: coverage and translation}
\label{app:mt}

A \rpt{} tokenizer routes analysable words through the root-and-pattern path. All other words go
through a fallback that is identical in the real and control arms. A real-minus-control contrast
therefore acts only on analysable words. For a test sentence $i$, let $c_i$ be the fraction of its
reference words that receive an analysis and $d_i$ the sentence-level score difference between the
real arm and a control.

\begin{theorem}[Coverage attenuation]
\label{thm:cov}
Assume (A1) the sentence score is an average of per-word contributions, (A2) unanalysed words
contribute zero to $d_i$ in expectation, and (A3) analysed words contribute a mean $\theta$ that
does not depend on sentence length or on which other words are analysed. Then
$\mathbb{E}[d_i\mid c_i]=\theta c_i$, and the overall effect is $\Delta=\theta\,\mathbb{E}[c]$.
\end{theorem}
\begin{proof}
By (A1), $\mathbb{E}[d_i\mid c_i]$ averages the expected contributions of the $n_i$ words. By (A2)
and (A3), the $c_in_i$ analysed words contribute $\theta$ each, whereas the rest contribute zero.
The average is therefore $\theta c_i$. Taking expectations over sentences gives
$\Delta=\theta\,\mathbb{E}[c]$.
\end{proof}

\noindent
At mean coverage $0.37$, the smallest detectable per-word effect is therefore $2.7$ times that at
full coverage, and a null at low coverage says little about $\theta$.

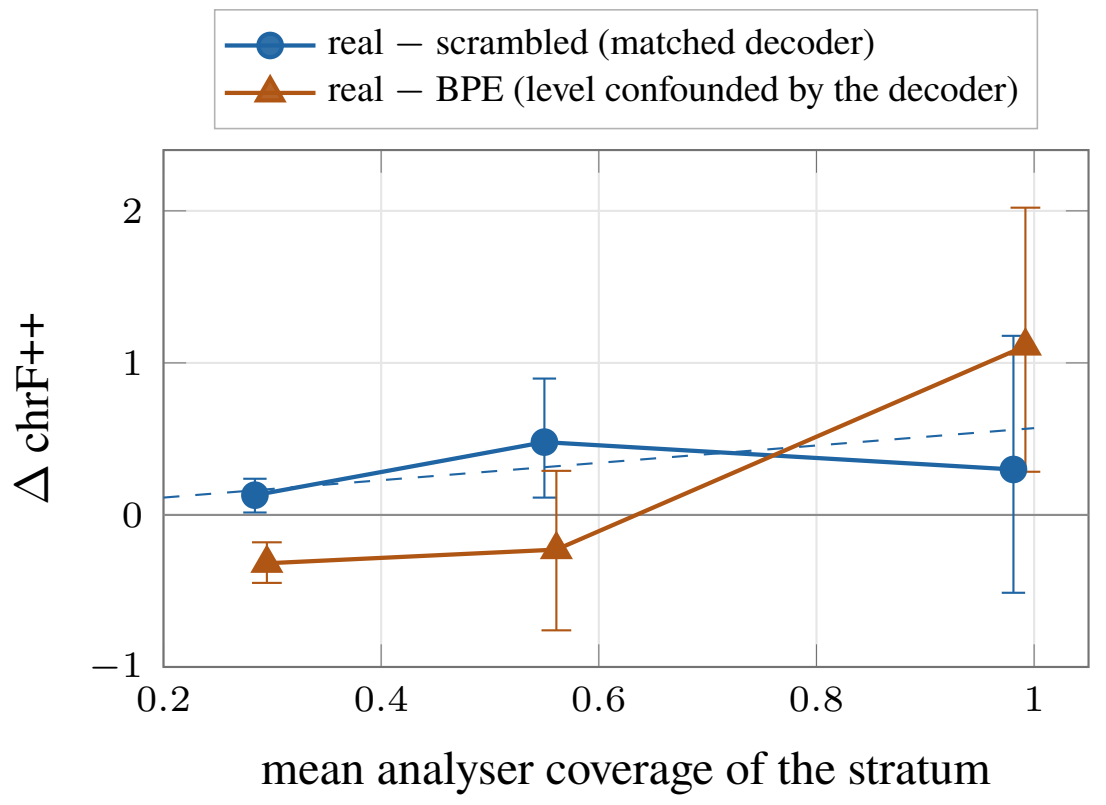
\begin{figure}[t]
\centering
\begin{tikzpicture}
\begin{axis}[
  paper, width=\linewidth, height=5.0cm,
  xlabel={mean analyser coverage of the stratum},
  ylabel={$\Delta$ \chrf{}},
  xmin=0.2, xmax=1.05, ymin=-1.0, ymax=2.4,
  grid=major,
  legend style={at={(0.5,1.04)},anchor=south},
  legend cell align=left,
]
\draw[black!45,line width=0.4pt] (axis cs:0.2,0) -- (axis cs:1.05,0);
\addplot+[tokcol,mark=*,mark size=2.2pt,mark options={fill=tokcol},thick,
  error bars/.cd, y dir=both, y explicit, error bar style={tokcol}]
  coordinates {(0.284,0.130) += (0,0.108) -= (0,0.114)
               (0.550,0.479) += (0,0.418) -= (0,0.365)
               (0.981,0.299) += (0,0.879) -= (0,0.811)};
\addlegendentry{real $-$ scrambled (matched decoder)}
\addplot+[enccol,mark=triangle*,mark size=2.8pt,mark options={fill=enccol},thick,
  error bars/.cd, y dir=both, y explicit, error bar style={enccol}]
  coordinates {(0.295,-0.318) += (0,0.138) -= (0,0.129)
               (0.561,-0.228) += (0,0.518) -= (0,0.531)
               (0.992,1.111) += (0,0.910) -= (0,0.827)};
\addlegendentry{real $-$ BPE (level confounded by the decoder)}
\addplot[tokcol,dashed,domain=0.2:1.0,samples=2,forget plot]{0.571*x};
\end{axis}
\end{tikzpicture}
\caption{Study~II en$\to$am translation by coverage stratum ($c<0.5$: $1{,}597$ sentences;
$0.5\le c<0.75$: $268$; $c\ge0.75$: $135$), with $95\%$ stratified-bootstrap intervals. Dashed
line: through-origin fit $\Delta=\theta_0c$ for the matched contrast. Triangles are shifted
slightly to the right.}
\label{fig:covmt}
\end{figure}

\paragraph{Study II set-up.} Study~II trains on $60{,}000$ en--am pairs with a $4.06$M-parameter
model and a target table of $10{,}000$ rows for every arm. Once roots are randomised, a root token
no longer identifies a word (Proposition~\ref{prop:ctlred}), and all \rpt{} arms are therefore
detokenised by one exact-match lexicon of training target types. This decoder matches real and
control arms exactly. However, it penalises the \rpt{} arms relative to BPE, and only the
dependence on coverage is interpretable for the contrasts between \rpt{} arms and BPE. All five
systems are weak ($7.2$--$8.2$ \chrf{}, BPE included) for undiagnosed reasons.

\paragraph{Results.} Coverage is skewed: $1{,}597$ of $2{,}000$ test sentences have $c<0.5$. Real
roots beat scrambled roots by $+0.19$ \chrf{} $[0.07, 0.31]$ over test sentences ($p=0.002$), and
the two permuted controls by $+0.29$ and $+0.62$ (Table~\ref{tab:covmt}). Across seeds, the
contrast with scrambling is not significant ($+0.31$, $\pseed=0.11$), whereas the contrasts with
the two permutations are positive in all five seeds ($\pseed=0.064$ and $0.001$). The
through-origin slope $\theta_0=0.57$ predicts an overall effect of $0.21$ against a measured $0.19$
for scrambling, and $0.28$ against a measured $0.29$ for the plain permutation. For the stratified
permutation the fit fails, which indicates that (A3) does not hold for that control. Against BPE,
the \rpt{} tokenizer trails below $c=0.5$ ($-0.32$) and leads above $c=0.75$ ($+1.11$). This
interaction of $+1.43$ $[0.58, 2.35]$ also appears for the scrambled arm ($+1.26$), which indicates
that it partly reflects granularity.

\begin{table*}[t]
\centering
\footnotesize
\setlength{\tabcolsep}{3pt}
\begin{tabular}{lccccccc}
\toprule
\textbf{Contrast} & $c<0.5$ & $0.5\le c<0.75$ & $c\ge0.75$ & \textbf{overall} & \textbf{high $-$ low} & $\theta_0$ [95\% CI] & $\theta_0\mathbb{E}[c]$ vs.\ $\Delta$ \\
\midrule
\arm{real} $-$ \arm{scrambled} & $+0.130^{*}$ & $+0.479^{*}$ & $+0.299$ & $+0.188^{*}$ & $+0.169$ & $0.571$ $[0.213,0.960]$ & $0.210$ vs.\ $0.188$ \\
\arm{real} $-$ \arm{perm}      & $+0.278^{*}$ & $+0.203$ & $+0.662$ & $+0.294^{*}$ & $+0.384$ & $0.751$ $[0.405,1.092]$ & $0.275$ vs.\ $0.294$ \\
\arm{real} $-$ \arm{permstrat} & $+0.671^{*}$ & $+0.759^{*}$ & $-0.223$ & $+0.622^{*}$ & $-0.894^{*}$ & $1.206$ $[0.879,1.536]$ & $0.443$ vs.\ $0.622$ \\
\arm{perm} $-$ \arm{scrambled} & $-0.148^{*}$ & $+0.276$ & $-0.362$ & $-0.105^{\mathrm{eq}}$ & $-0.215$ & -- & -- \\
\arm{real} $-$ \arm{bpe}$^{\dagger}$      & $-0.318^{*}$ & $-0.228$ & $+1.111^{*}$ & $-0.210^{*}$ & $+1.429^{*}$ & -- & -- \\
\arm{scrambled} $-$ \arm{bpe}$^{\dagger}$ & $-0.448^{*}$ & $-0.707^{*}$ & $+0.811^{*}$ & $-0.398^{*}$ & $+1.260^{*}$ & -- & -- \\
\bottomrule
\end{tabular}
\caption{Study~II coverage-stratified MT (en$\to$am, five seeds, sentence-level \chrf{}, stratified
paired bootstrap). $\theta_0$: through-origin coverage slope; the last column compares
$\theta_0\mathbb{E}[c]$ with the measured overall effect. Mean corpus \chrf{} over seeds: \arm{bpe}
$8.20$, \arm{real} $8.05$, \arm{scrambled} $7.73$, \arm{perm} $7.60$, \arm{permstrat} $7.23$.
$^{*}$\,$95\%$ interval excludes zero. $^{\mathrm{eq}}$\,equivalent to zero at $\delta=0.5$.
$^{\dagger}$\,level confounded by the shared lexicon decoder.}
\label{tab:covmt}
\end{table*}

\section{Experimental details}
\label{app:details}

\paragraph{Study I hyperparameters.}\label{tab:hparams} \emph{NER:} pre-LN Transformer encoder with
2 layers, $d=128$, 4 heads, FFN 256 and dropout $0.1$ ($965$k--$967$k parameters, $275{,}601$
non-embedding); AdamW (weight decay $0.01$), learning rate $3\times10^{-4}$ with $10\%$ warm-up and
linear decay, batch 16, 12 epochs, clipping $1.0$; 48-word sequence budget. \emph{MT:} pre-LN
encoder--decoder with 2+2 layers, $d=256$, 4 heads and FFN 512 ($8.84$M parameters for Amharic,
$7.47$M for Hausa); AdamW, learning rate $5\times10^{-4}$, $5\%$ warm-up, batch 32, 8 epochs, label
smoothing $0.1$; 40 source and 40 target words; greedy decoding. Seeds are $13$, $21$, $34$, $55$
and $89$. Table~\ref{tab:studyii} lists the changes made in Study~II.

\begin{table*}[t]
\centering
\footnotesize
\setlength{\tabcolsep}{5pt}
\begin{tabular}{>{\raggedright\arraybackslash}p{0.13\linewidth}>{\raggedright\arraybackslash}p{0.38\linewidth}>{\raggedright\arraybackslash}p{0.41\linewidth}}
\toprule
& \textbf{Study I} ($435$ runs) & \textbf{Study II} ($161$ runs) \\
\midrule
Root source & HornMorpho analyses, lexicon, LCS induction & LCS induction only; $1{,}238$ roots of $3$--$4$ radicals \\
Coverage; $\fert$ & $80.4\%$ of NER types; $\fert=2.82$--$2.90$ (BPE $1.62$) & $67.2\%$ of NER types (MT: $60.6\%$); $\fert=2.17$ (BPE $1.53$) \\
Controls & scrambled; fertility-matched BPE; granularity sweep & scrambled; permuted; stratified permutation \\
NER model & $0.97$M parameters, 12 epochs, final epoch & $1.95$M parameters, $\le30$ epochs, best development epoch (patience 6) \\
MT & $20$k pairs, $8.8$M parameters, exact inversion & $60$k pairs, $4.06$M parameters, shared lexicon decoder \\
Pretrained & none & XLM-R base, AfroXLM-R base; 3 seeds \\
Channel $F$ & gemination and root class & root-class stand-in ($8$ classes) \\
\bottomrule
\end{tabular}
\caption{What Study~II changes relative to Study~I.}
\label{tab:studyii}
\end{table*}

\paragraph{Capacity matching in Study II.} Every Study~II NER model uses an embedding table of
$12{,}000$ rows, and all arms, BPE included, therefore have $1{,}954{,}441$ parameters
($1{,}955{,}721$ with an encoder channel). The number of token types that occur in the training
text differs slightly between the \rpt{} arms ($6{,}277$ to $6{,}298$ types, against $5{,}115$ for
BPE), because roots are reassigned over the word types of all three splits and a reassignment
changes which root identifiers occur in training. The fertility of the four \rpt{} arms is
identical ($2.1725$).

\paragraph{Data and compute.} The OPUS-100 filter reduced the Amharic corpus from $89{,}027$ to
$52{,}564$ pairs, mainly by length, and the Hausa corpus from $97{,}983$ to $87{,}583$ pairs.
Study~I comprises $435$ runs and took $338.7$ minutes, plus a one-off HornMorpho analysis of about
$5.7$ hours. Study~II comprises $161$ runs ($100$ NER, $25$ MT, $36$ fine-tuning) and took about
$265$ minutes. Corpora are retrieved with recorded SHA-256 hashes. Correctness gates (the round
trip of Theorem~\ref{thm:lossless}; identical fertility and identical multisets of root identifiers
across real and permuted arms; equal non-embedding parameters) halt a run on failure.

\section{Additional results}
\label{app:results}

\subsection{Study I}
\label{app:study1}

Table~\ref{tab:readout1} gives the full Study~I NER results. All Amharic readout gains $\Gamma$ and
all pooled comparisons with BPE have $\pholm\le0.006$, and all first-sub-token comparisons have
$\pholm=1.0$. For Hausa (first/pool), \arm{bpe} scores $48.01/47.90$ with $\Gamma=-0.10$
$[-0.87,0.64]$, \arm{unigram} $48.04/43.93$ with $\Gamma=-4.11$ $[-5.48,-2.73]$, and
fertility-matched BPE $42.11/39.13$ with $\Gamma=-2.98$ $[-4.90,-1.14]$.

\begin{table*}[t]
\centering
\footnotesize
\setlength{\tabcolsep}{2.5pt}
\begin{tabular}{lcccccc}
\toprule
\textbf{Arm} & \textbf{First} & \textbf{Root token} & \textbf{Pool} & $\Gamma$ [95\% CI] & \textbf{vs.\ \arm{bpe}, first} & \textbf{vs.\ \arm{bpe}, pool} \\
\midrule
\arm{bpe}      & $14.73\pm1.18$ & --              & $16.74\pm1.49$ & $+2.01$ $[1.02,3.03]$ & -- & -- \\
\arm{unigram}  & $15.43\pm0.94$ & --              & $19.36\pm1.39$ & $+3.93$ $[2.24,5.63]$ & $+0.70$ $[-0.94,2.34]$ & $+2.62$ $[1.02,4.35]$ \\
\arm{lexicon}  & $14.26\pm1.79$ & $25.33\pm0.82$ & $25.49\pm1.12$ & $+11.23$ $[8.47,14.06]$ & $-0.47$ $[-3.14,2.07]$ & $+8.75$ $[6.27,11.30]$ \\
\arm{analyser} & $12.47\pm0.58$ & $23.69\pm0.66$ & $25.85\pm1.94$ & $+13.38$ $[10.74,16.08]$ & $-2.26$ $[-4.32,-0.15]$ & $+9.11$ $[6.58,11.71]$ \\
\arm{induced}  & $12.24\pm1.42$ & $21.71\pm1.40$ & $23.81\pm0.49$ & $+11.57$ $[8.62,14.49]$ & $-2.50$ $[-4.78,-0.10]$ & $+7.07$ $[4.66,9.49]$ \\
\bottomrule
\end{tabular}
\caption{Study~I Amharic NER by tokenizer and readout (F1 $\times100$, mean $\pm$ between-seed SD).
``Root token'' reads the word label from the state of the root token.}
\label{tab:readout1}
\end{table*}

\paragraph{Study I translation.}\label{tab:mt1} Every arm with substantially higher target
fertility than BPE scores lower. One possible reason is that a decoder must emit every extra token,
although our design does not isolate this mechanism. Corpus \chrf{} (five seeds, between-seed SD
$0.20$; target fertility in parentheses) is as follows. Amharic: \arm{unigram} $17.32$ ($1.41$), \arm{bpe} $16.91$ ($1.25$),
\arm{scrambled} $16.01$ ($2.60$), \arm{analyser} $15.96$ ($2.60$), \arm{induced} $15.81$ ($2.55$)
and \arm{lexicon} $15.44$ ($2.52$). Hausa: \arm{bpe} $31.09$ ($1.27$), \arm{unigram} $30.63$
($1.47$) and fertility-matched BPE $29.57$ ($1.97$). Every difference from \arm{bpe} has
$\pholm\le0.003$.

\subsection{Study II}
\label{app:study2}

\begin{table*}[t]
\centering
\footnotesize
\setlength{\tabcolsep}{6pt}
\begin{tabular}{lcccccl}
\toprule
\textbf{Contrast} & $\Delta$ & \textbf{95\% CI} & $\pholm$ & $\pseed$ ($n_{+}$) & \textbf{Verdict} \\
\midrule
$\Dg$: \arm{scrambled} $-$ \arm{bpe}         & $-3.39$ & $[-5.34,-1.50]$ & $0.012$ & $0.006$ (0) & effect (negative) \\
$\Dr$: \arm{real} $-$ \arm{scrambled}        & $+4.65$ & $[+3.01,+6.33]$ & $0.006$ & $0.011$ (5) & effect \\
\arm{real} $-$ \arm{perm}                    & $+2.92$ & $[+1.32,+4.49]$ & $0.006$ & $0.16$ (4) & effect (sentences only) \\
\arm{real} $-$ \arm{permstrat}               & $+5.21$ & $[+3.65,+6.76]$ & $0.006$ & $0.055$ (4) & effect (sentences only) \\
residual: \arm{perm} $-$ \arm{scrambled}     & $+1.73$ & $[+0.19,+3.32]$ & $0.070$ & $0.14$ (5) & inconclusive \\
residual: \arm{permstrat} $-$ \arm{scrambled} & $-0.56$ & $[-2.09,+0.99]$ & $0.490$ & $0.72$ (1) & equivalent at 2 \\
total: \arm{real} $-$ \arm{bpe}              & $+1.25$ & $[-0.47,+3.04]$ & $0.154$ & $0.48$ (3) & inconclusive \\
\bottomrule
\end{tabular}
\caption{Study~II identification family (pooled readout, F1 $\times100$). $\pseed$: paired $t$-test
over seeds; $n_{+}$: number of seeds with a positive difference; verdicts at $\delta=2$ F1.}
\label{tab:ident2}
\end{table*}

\begin{table*}[t]
\centering
\footnotesize
\setlength{\tabcolsep}{5pt}
\begin{tabular}{llcccl}
\toprule
\textbf{Model} & \textbf{Cell} & \textbf{Per-seed F1} & \textbf{F1} & \textbf{SD} & \textbf{Contrast and verdict} \\
\midrule
\multirow{4}{*}{XLM-R}
 & first, none & $66.16/70.72/70.52$ & $69.11$ & $2.58$ & pool $-$ first: $+0.72$ $[-1.53,+2.95]$, inconclusive \\
 & pool, none  & $71.47/66.73/0.00^{\ddagger}$ & $69.11^{\S}$ & $3.35^{\S}$ & -- \\
 & pool, $F$   & $67.79/69.82/70.53$ & $69.37$ & $1.42$ & $F-$none: $-0.32$, equivalent \\
 & pool, $D$   & $70.06/67.07/67.77$ & $68.29$ & $1.56$ & $D-$none: $-0.54$, inconclusive \\
\midrule
\multirow{4}{*}{AfroXLM-R}
 & first, none & $73.10/75.43/74.45$ & $74.32$ & $1.17$ & pool $-$ first: $+0.54$ $[-1.21,+2.10]$, equivalent \\
 & pool, none  & $73.69/75.71/75.19$ & $74.86$ & $1.05$ & -- \\
 & pool, $F$   & $73.22/74.39/75.19$ & $74.27$ & $0.99$ & $F-$none: $-0.60$ $[-1.75,+0.57]$, equivalent \\
 & pool, $D$   & $73.67/73.32/74.25$ & $73.75$ & $0.47$ & $D-$none: $-1.12$ $[-2.26,+0.03]$, inconclusive \\
\bottomrule
\end{tabular}
\caption{Pretrained encoders on Amharic NER (three seeds, fertility $2.20$). F1 is computed from
per-sentence test counts averaged over seeds, as for every Study~II cell. $^{\ddagger}$\,divergent
run (development F1 also zero), rejected by the development-score gate. $^{\S}$\,over the two
convergent seeds (over all three: count-based F1 $55.59$, SD $39.96$). Channel cells under
first-sub-token readout ($69.67$, $68.90$ for XLM-R; $74.87$, $74.12$ for AfroXLM-R) are omitted.
Verdicts at $\delta=2$ F1.}
\label{tab:pretrained}
\end{table*}

\paragraph{A divergent run.} One of the $36$ fine-tuning runs (XLM-R, pooled readout, third seed)
never left its initialisation and scored zero F1 on both development and test data. Had it been
included, the sentence-level bootstrap would have reported a readout effect of $-13.5$ F1
($\pholm=0.001$), as large as the largest genuine effect in the study. The development-score gate
removed this run, and no other run triggered the gate. The reported XLM-R contrasts use the two
convergent seeds.

\paragraph{Control diagnostics.} The permutation controls behave as intended. The largest root
identifier holds $168$ word types in both the real and the permuted arms, against $21$ under
scrambling. The rank correlation of per-identifier token frequencies with the real arm rises from
$-0.03$ (scrambled) to $0.79$ (permuted) and $0.97$ (stratified).

\paragraph{Per-seed scores.}\label{tab:seeds} Study~II pooled-readout F1 for seeds 0 to 4:
\arm{bpe} $22.37$, $22.85$, $24.03$, $23.23$, $21.61$; \arm{real} $22.85$, $21.28$, $23.72$,
$23.29$, $29.39$; \arm{scrambled} $18.70$, $18.85$, $19.80$, $19.06$, $20.68$; \arm{perm} $19.59$,
$24.07$, $20.00$, $20.57$, $21.30$; \arm{permstrat} $17.28$, $22.16$, $19.62$, $17.71$, $18.42$.

\paragraph{Equivalence verdicts.} Nine nulls are equivalences at the pre-specified margins. Eight
of them use $\delta=2$ F1: the frequency residual of the stratified permutation, the non-derivable
channel behind BPE and behind real roots, two interactions for the permuted arm (with the readout
and with the derivable channel), the AfroXLM-R readout gain, and the non-derivable channel in both
pretrained encoders. The ninth is the MT difference between the permuted and scrambled controls
($\delta=0.5$ \chrf{}). All other nulls are only inconclusive.

\end{document}